\documentclass{article}
\usepackage{authblk}
\usepackage{cite}
\usepackage{graphicx}
\usepackage{subcaption}
\usepackage[table]{xcolor}
\usepackage{enumerate}
\usepackage{mathpazo}
\usepackage[utf8]{inputenc}

\providecommand{\keywords}[1]
{
  \small	
  \textbf{\textit{Keywords---}} #1
}

\definecolor{red}{RGB}{255,0,0.}
\definecolor{yellow}{RGB}{218,165,32}
\title{Automated Diagnosis of Intestinal Parasites: A new hybrid approach and its benefits}
\author[1,a]{Daniel Osaku}
\author[2,a]{C. F. Cuba}
\author[3,a]{Celso T.N. Suzuki}
\author[4,a]{J.F. Gomes}
\author[5,a]{A.X. Falc\~{a}o}
\affil{danosaku@hotmail.com, $^2$carolinacuba23@gmail.com, $^3$celso.suzuki@gmail.com, $^4$jgomes@ic.unicamp.br, $^5$afalcao@ic.unicamp.br}
\affil[a]{Institute of Computing, University of Campinas, Brazil}
\usepackage{multirow, multicol}
\usepackage{algorithm}
\usepackage{algorithmicx}
\usepackage{algpseudocode}
\usepackage{url}
\begin{document}

\maketitle

\begin{abstract}
\begin{sloppypar}

  Intestinal parasites are responsible for several diseases in human
  beings. In order to eliminate the error-prone visual analysis of
  optical microscopy slides, we have investigated automated, fast, and
  low-cost systems for the diagnosis of human intestinal parasites. In
  this work, we present a hybrid approach that combines the opinion of
  two decision-making systems with complementary properties: ($DS_1$)
  a simpler system based on very fast handcrafted image feature
  extraction and support vector machine classification and ($DS_2$) a
  more complex system based on a deep neural network, Vgg-16, for
  image feature extraction and classification. $DS_1$ is much faster
  than $DS_2$, but it is less accurate than $DS_2$. Fortunately, the
  errors of $DS_1$ are not the same of $DS_2$. During training, we use
  a validation set to learn the probabilities of misclassification by
  $DS_1$ on each class based on its confidence values. When $DS_1$
  quickly classifies all images from a microscopy slide, the method
  selects a number of images with higher chances of misclassification
  for characterization and reclassification by $DS_2$. Our hybrid
  system can improve the overall effectiveness without compromising
  efficiency, being suitable for the clinical routine --- a strategy
  that might be suitable for other real applications. As demonstrated on
  large datasets, the proposed system can achieve, on average, 94.9\%,
  87.8\%, and 92.5\% of Cohen's Kappa on helminth eggs, helminth
  larvae, and protozoa cysts, respectively.
  \end{sloppypar}

\end{abstract}


\section{Introduction}

A recent report by the World Health Organization indicates that
approximately 1.5 billion people are infected with intestinal
parasites~\cite{WHO:2018}. These parasitic diseases are most common in tropical countries due to climate, precarious health services, poor sanitary
conditions, among several other factors. The
problem can cause mental and physical disorders (e.g., the difficulty of
concentration, diarrhea, abdominal pain) or, in the extreme case,
death, especially in infants and immunodeficient individuals.

\begin{figure}[!ht]
  \begin{center}
    \begin{tabular}{cccccc}
      \includegraphics[width=.1197\linewidth]{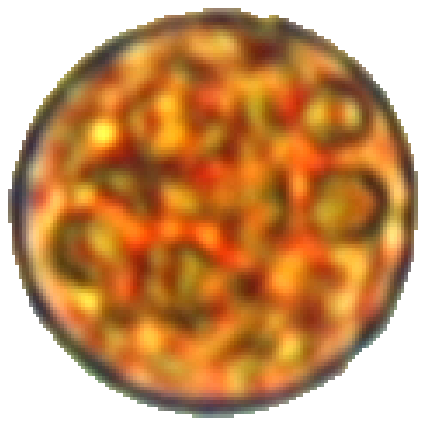} &
      \includegraphics[width=.057285\linewidth]{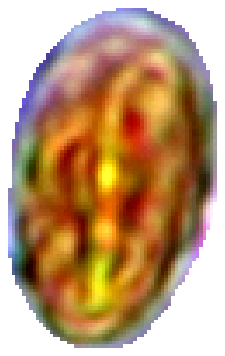} &
      \includegraphics[width=.171\linewidth]{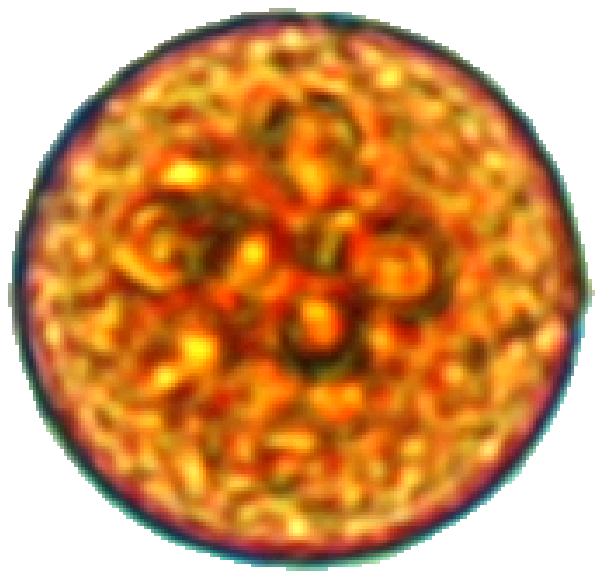} &
      \includegraphics[width=.05985\linewidth]{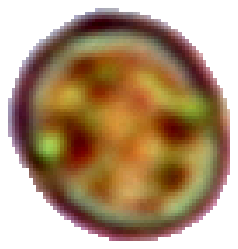} &
      \includegraphics[width=.0855\linewidth]{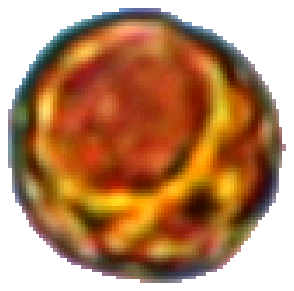} &
      \includegraphics[width=.074385\linewidth]{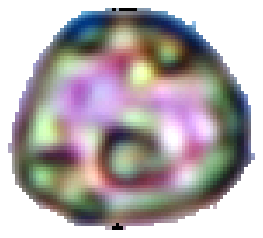} \\
      (a) &
      (b) &
      (c) &
      (d) &
      (e) &
      (f) \\
    \end{tabular}
  \end{center}
  \caption{Examples of protozoan.
    (a) \emph{Entamoeba histolytica}/\emph{E. dispar};
    (b) \emph{Giardia duodenalis};
    (c) \emph{Entamoeba coli};
    (d) \emph{Endolimax nana};
    (e) \emph{Iodamoeba b\"utschlii};
    (f) \emph{Blastocystis hominis}.}
  \label{f.parasitos-proto}
\end{figure}

\begin{figure}[ht]
  \begin{center}
    \begin{tabular}{cccc}
      \includegraphics[width=.12920\linewidth]{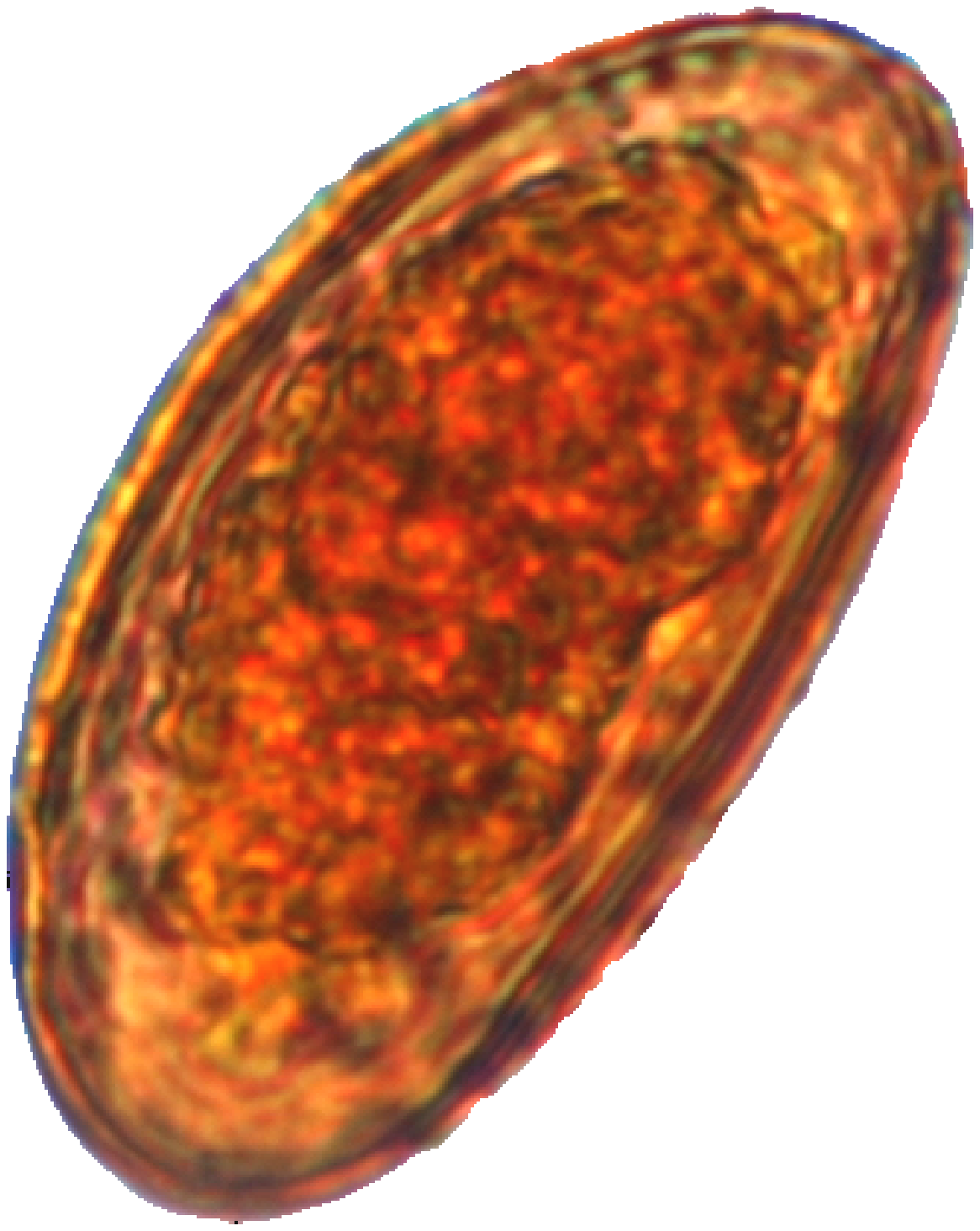} &
      \includegraphics[width=.142120\linewidth]{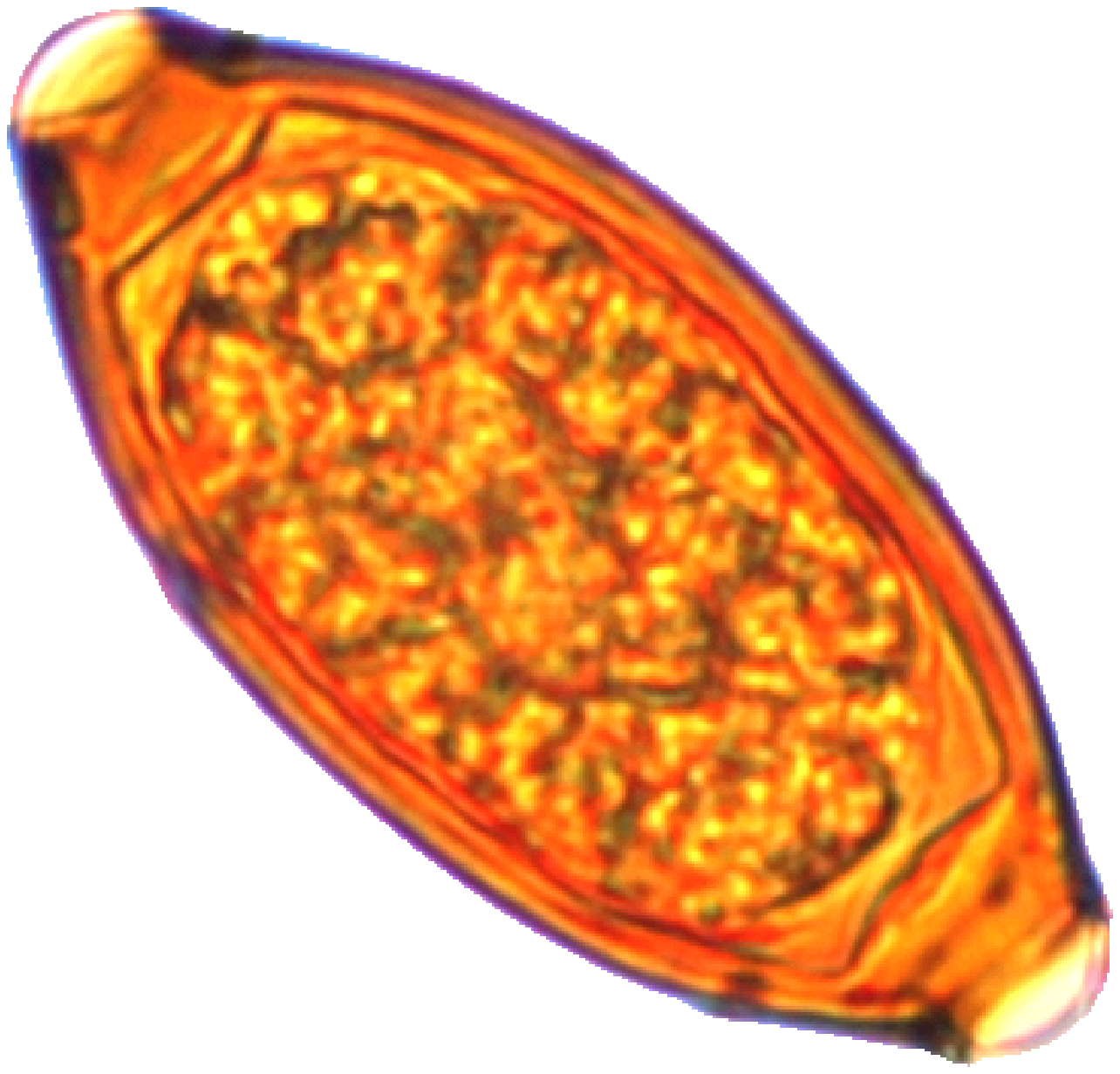} &
      \includegraphics[width=.134520\linewidth]{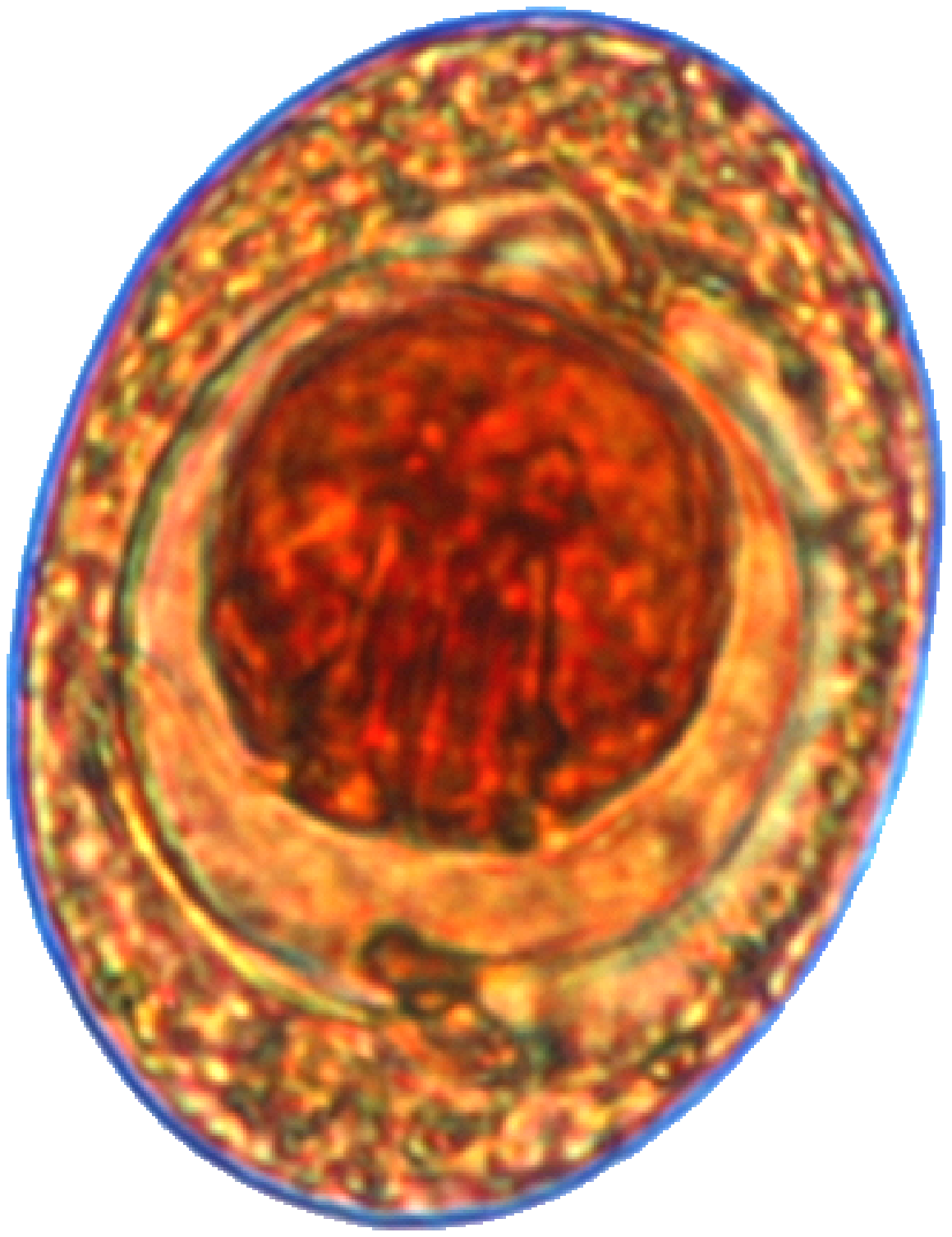} &
      \includegraphics[width=.119320\linewidth]{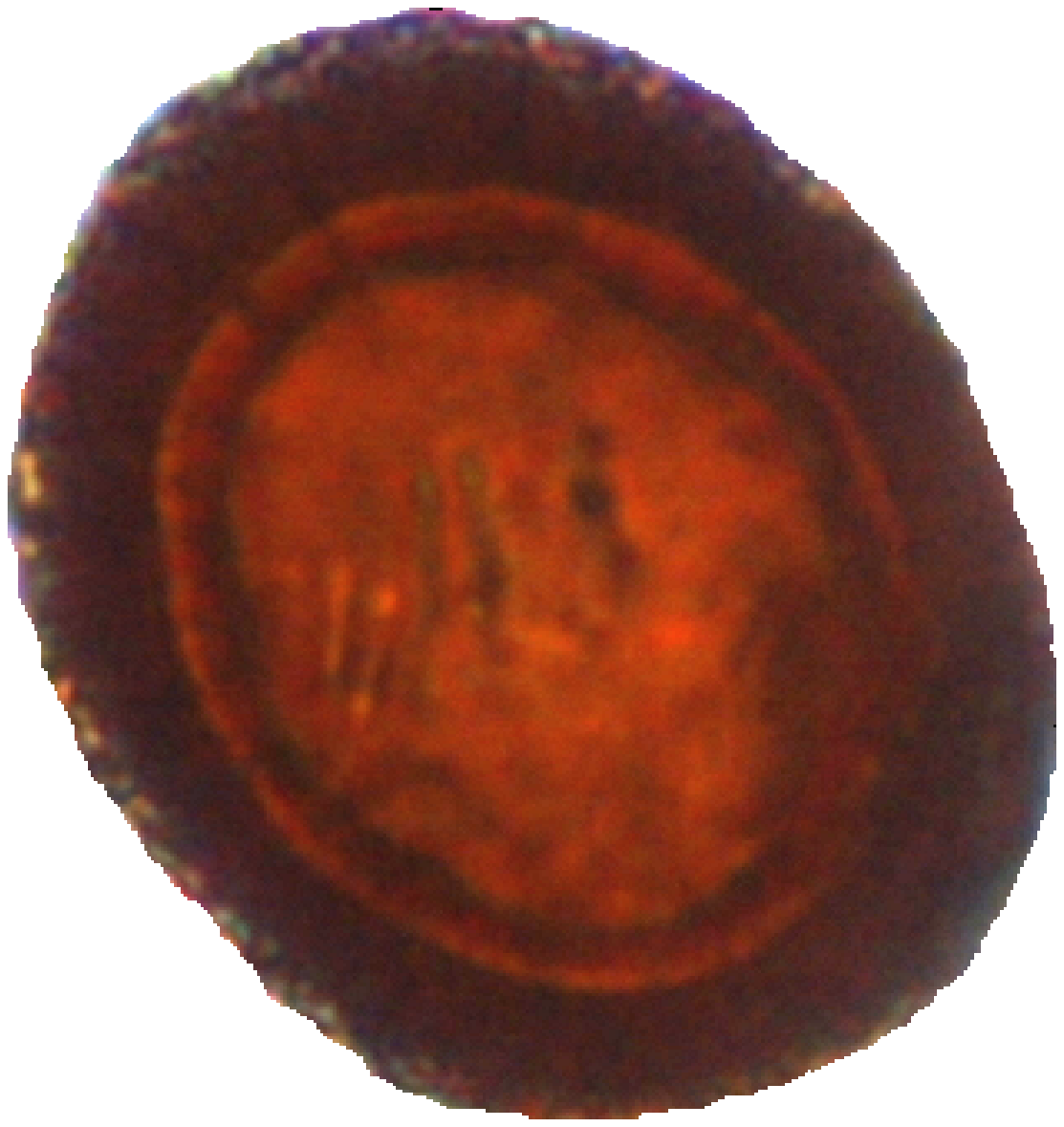} \\
      (a) &
      (b) &
      (c) &
      (d) \\
      \multicolumn{4}{c}{
        \begin{tabular}{ccc}
          \includegraphics[width=.2160\linewidth]{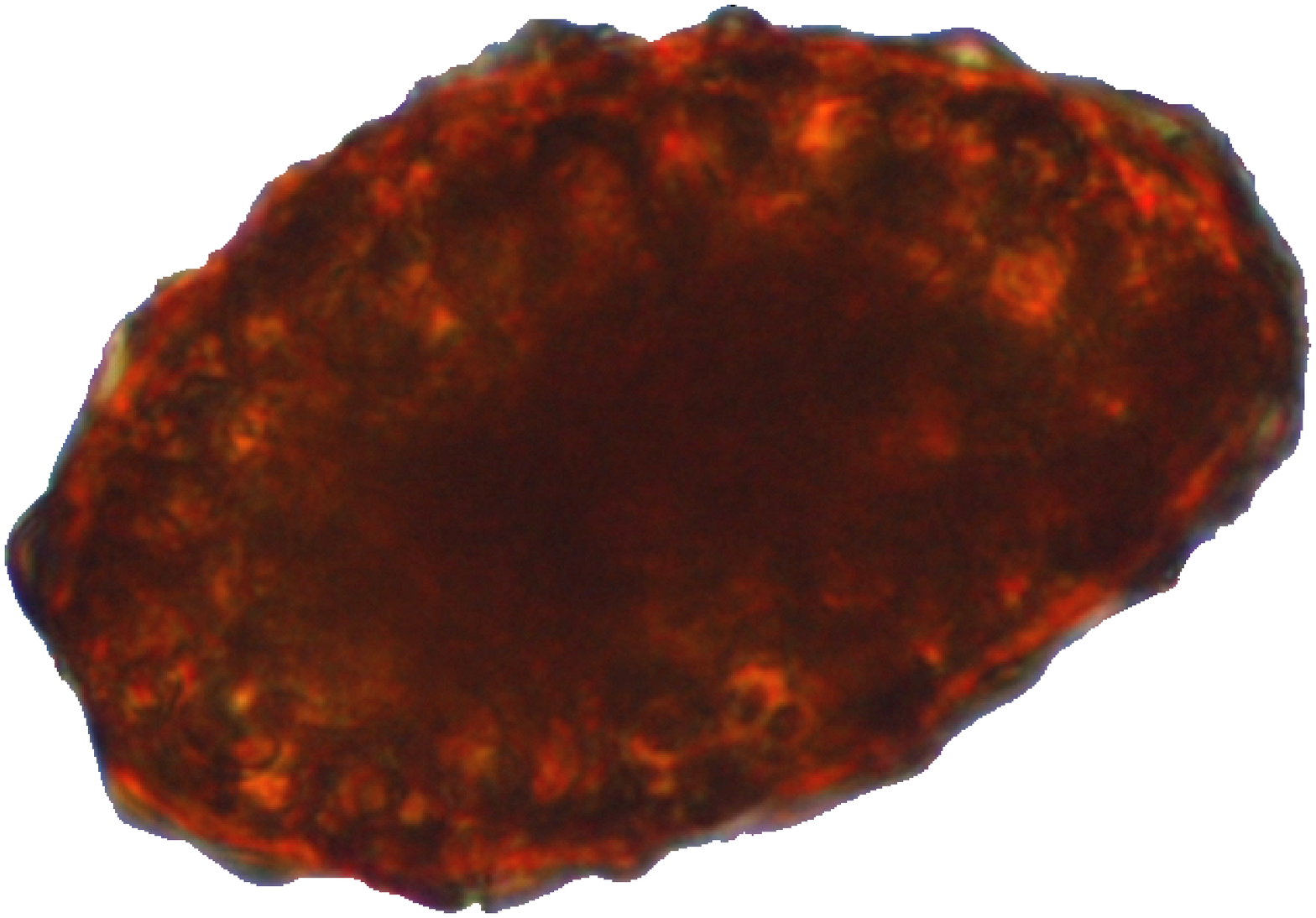} &
          \includegraphics[width=.1440\linewidth]{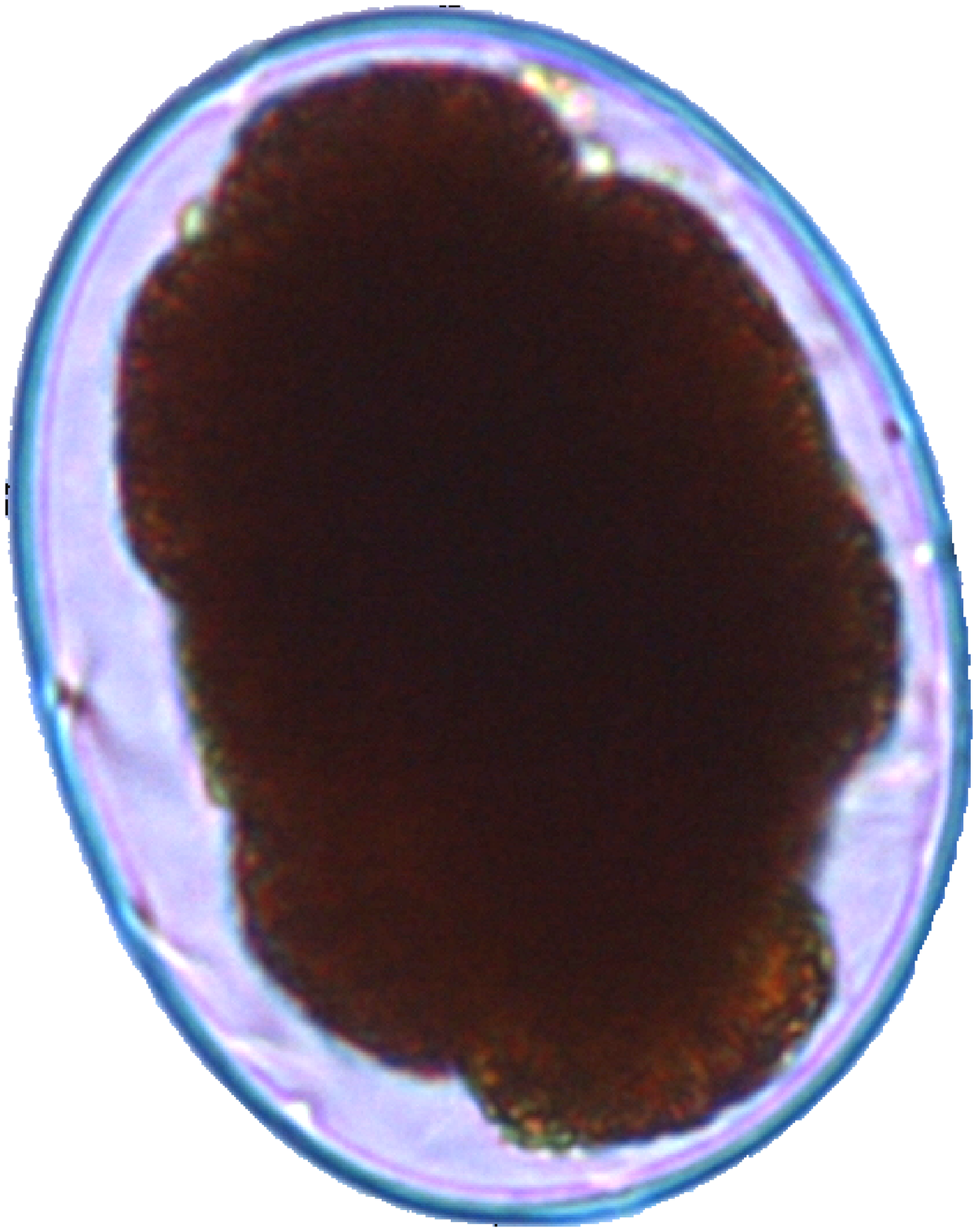} &
          \includegraphics[width=0.2120\linewidth]{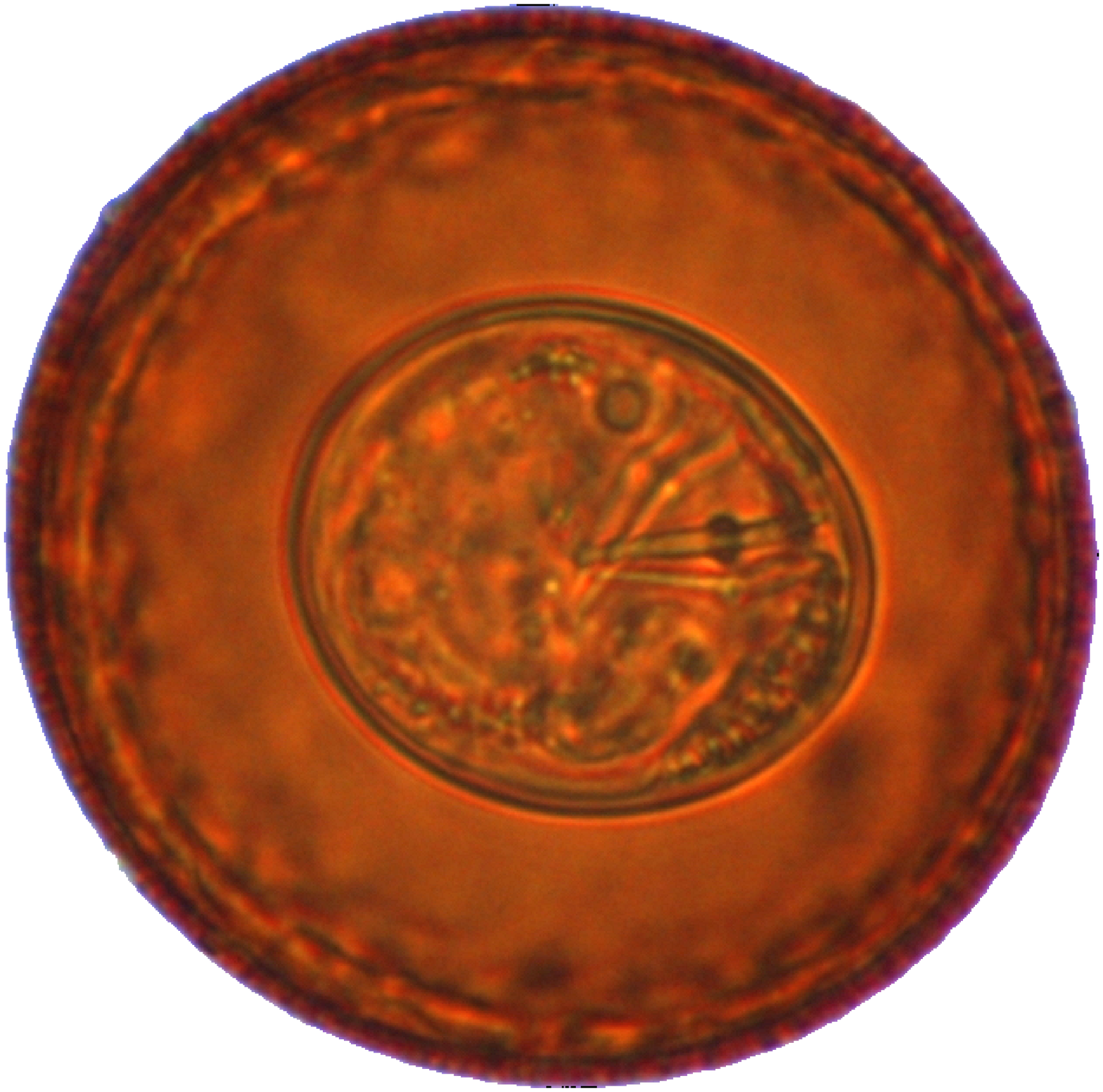} \\
          (e) &
          (f) &
          (g) \\
        \end{tabular}
      } \\
      \multicolumn{4}{c}{
        \begin{tabular}{cc}
          \includegraphics[width=0.32\linewidth]{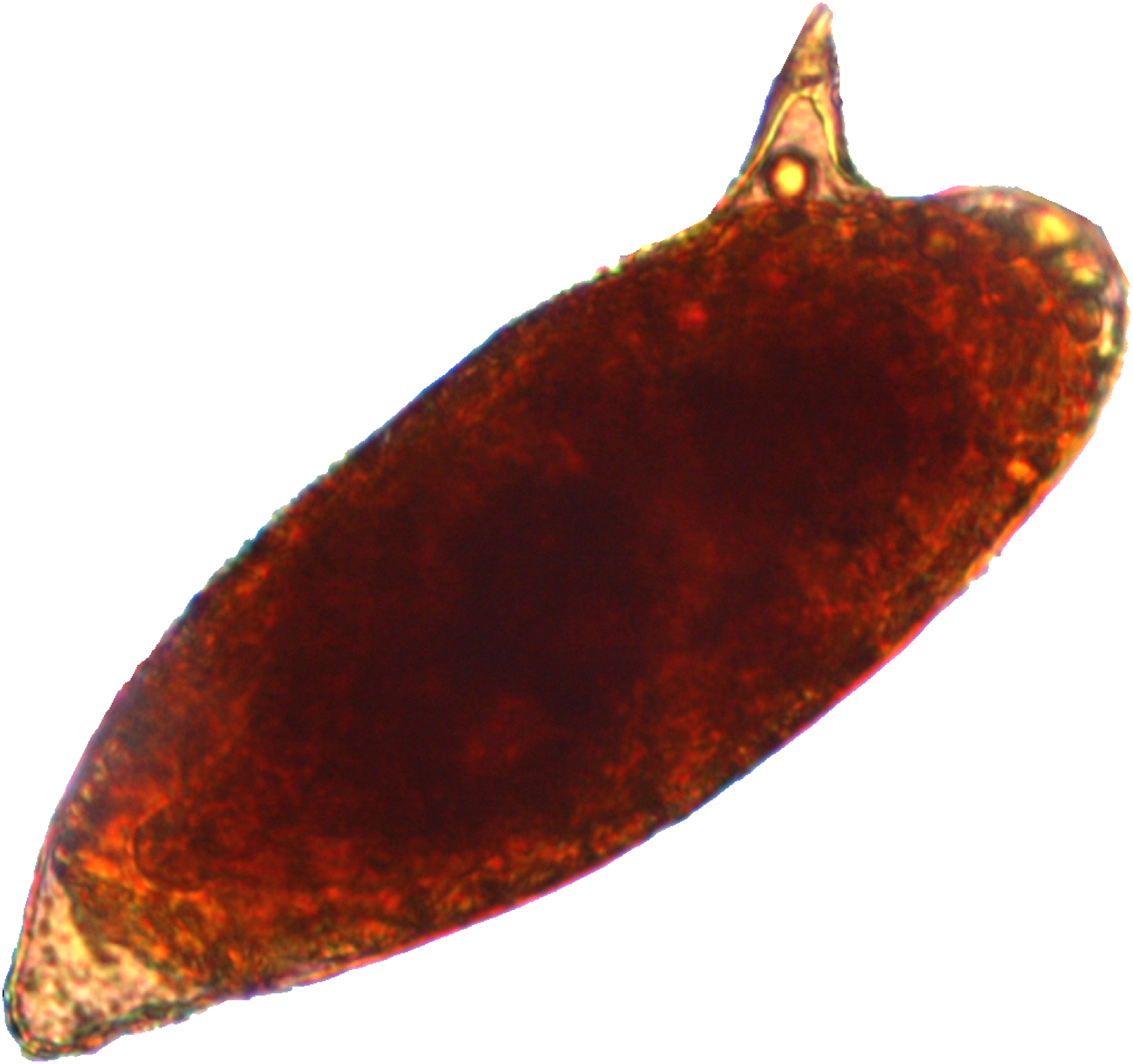} &
          \includegraphics[width=0.256\linewidth]{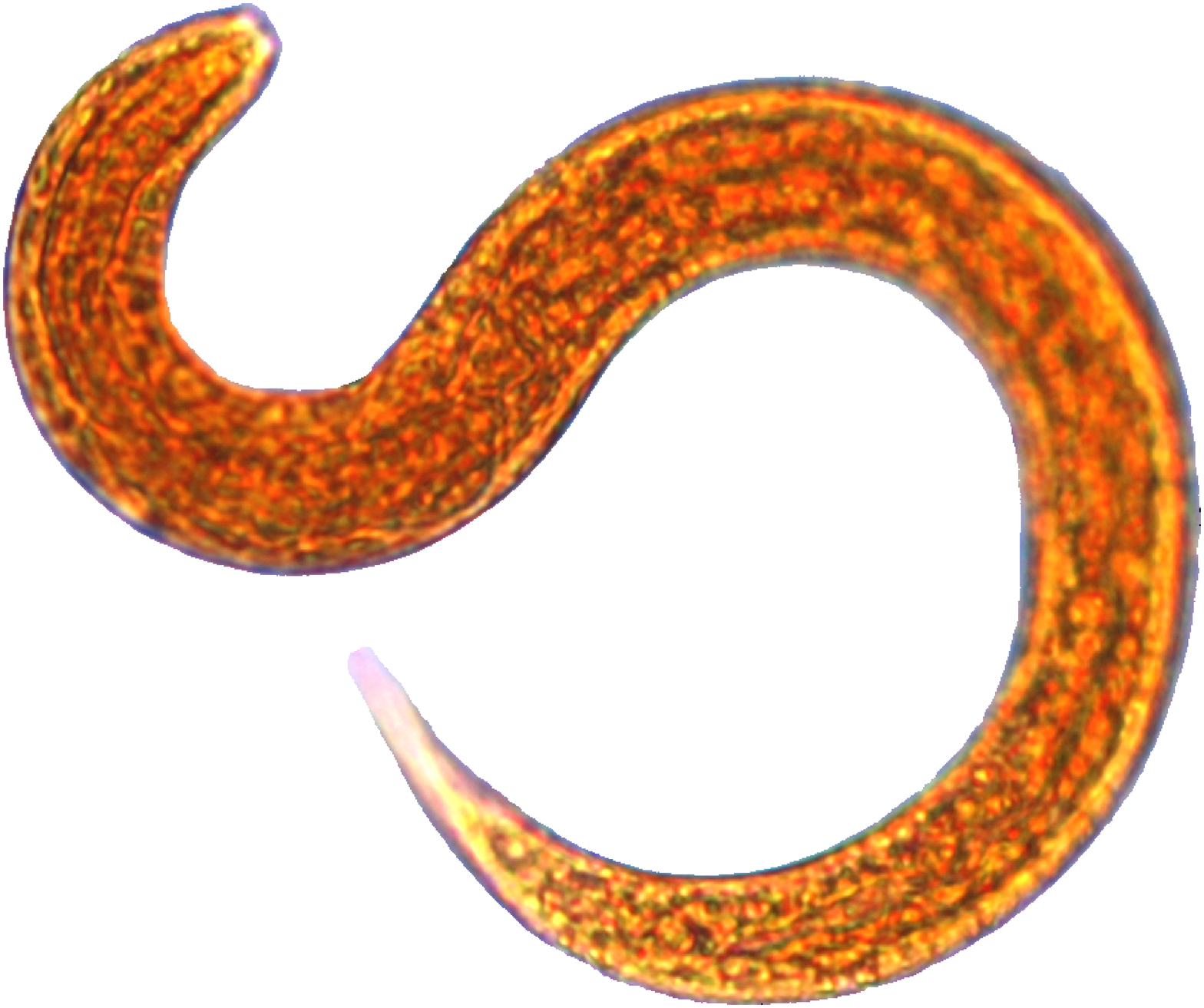} \\
          (h) &
          (i) \\
        \end{tabular}
      }
    \end{tabular}
  \end{center}
  \caption{Examples of helminths:
    (a) \emph{Enterobius vermicularis};
    (b) \emph{Trichuris trichiura};
    (c) \emph{Hymenolepis nana};
    (d) \emph{Taenia} spp.;
    (e) \emph{Ascaris lumbricoides};
    (f) Ancylostomatidae;
    (g) \emph{Hymenolepis diminuta};
    (h) \emph{Schistosoma mansoni};
    (i) \emph{Strongyloides stercoralis} larvae.}
  \label{f.parasitos-helm}
\end{figure}

The diagnostic procedure of the causative agent for parasitic
infections still relies on the visual analysis of optical microscopy
slides --- an error-prone procedure that usually results in low to
moderate diagnostic sensitivity~\cite{Carvalho2016TFTestMN}. In order
to circumvent the problem, we have developed the first automated
system for the diagnosis (the DAPI system) of the 15 most common
species of human intestinal parasites in
Brazil~\cite{Suzuki:2013:TBME,Suzuki:2013:ISBI}. Examples are
presented in Figures~\ref{f.parasitos-proto}
and~\ref{f.parasitos-helm}.

The DAPI system can produce about 2,000 images per microscopy slide,
with 4M pixels each and 12 bits per color channel. These images are
acquired on a compromise focus plane~\footnote{The parasites might
  appear at different focus depth, but it is impractical to find the
  optimum focus for each position of a microscopy slide.} and
processed in less than 4 minutes on a modern PC (Core i7 CPU with 16
threads) --- a time acceptable for the clinical routine. Our system
can successfully segment objects (parasites and similar impurities),
separate them into three groups, and align them for feature extraction
and classification. These groups are: (a) helminth eggs, (b) protozoa
cysts and vacuolar form of protozoa (Blastocystis hominis), and (c)
helminth larvae, all of them with similar fecal impurities which are
not eliminated during object segmentation.

The main question we seek to answer here is: can we
  benefit from the higher effectiveness of deep neural networks
  without compromising the efficiency and cost of the DAPI system?
Although machines have become faster, digital cameras can also
generate higher resolution images, and the DAPI system can also
improve and produce more images from multiple microscopy slides and
with different focus depth. Therefore, we believe that it will be
always desirable to improve effectiveness without compromising the
efficiency and cost of image analysis systems. Such a constraint is crucial to make the DAPI system viable for the
 public health system.

\begin{sloppypar}

In this work, we present a hybrid approach that combines the opinion
of two decision-making systems with complementary properties and
validate it for the diagnosis of those 15 most common species of human
intestinal parasites in Brazil. The first system ($DS_1$) is based on
very fast handcrafted image feature extraction~\cite{Suzuki:2013:ISBI}
and support vector machine (the probabilistic
$p$-SVM~\cite{wu2004probability}) classification. The time to extract
those handcrafted features is negligible and the $p$-SVM classifier
takes time equivalent to a decision layer of a deep neural network
since both can benefit from parallel matrix multiplication. As we
demonstrate in our experiments, among several SVM and deep neural
network models, we found $p$-SVM and Vgg-16~\cite{vggNet} the best
options for $DS_1$ and $DS_2$, respectively. Vgg-16 uses 15 extra
neuronal layers with a total number of neurons much higher than its
decision layer, which makes $DS_1$ about 30 times faster than $DS_2$
when using a same GPU board (GeForce GTX 1060 6GB/PCIe/SSE2). On the
other hand, $DS_2$ can be considerably more accurate than
$DS_1$. Fortunately, the errors of $DS_1$ and $DS_2$ are not the same,
given that they are statistically independent.
\end{sloppypar}

We then propose a hybrid approach that relies on a validation set to
learn the probabilities of misclassification by $DS_1$ on each class
based on its confidence values. Note that, as we will show, the
confidence values of $p$-SVM are not inversely proportional to its
chances of error. When $DS_1$ quickly classifies all images from a
microscopy slide, the method selects the images with higher chances to
have been misclassified by $DS_1$ and allows those images to be
characterized and reclassified by $DS_2$. By that, our method can
improve the overall effectiveness of the DAPI system without
compromising its efficiency and cost --- a strategy that can be used
in other real applications involving image analysis. Our experiments
demonstrate this result for large datasets by first comparing $DS_1$,
$DS_2$, and our hybrid approach. We then show a comparison among
several deep neural networks, that justifies our choice for Vgg-16,
and comparison among SVM models and the previous Optimum-Path Forest
classifier used in~\cite{Suzuki:2013:ISBI}, that justifies our choice
for $p$-SVM.

The remaining sections are organized as
follows. Section~\ref{s.relatedworks} presents the related works on
image classification of intestinal parasites and our previous version
of the automated system for the diagnosis of intestinal
parasites. Sections~\ref{s.materials} and~\ref{s.methods} present the
materials and the proposed hybrid approach based on $DS_1$ and
$DS_2$. The experimental results with discussion and the conclusions
are presented in Sections~\ref{s.experiments} and~\ref{s.conclusions},
respectively.

\section{Related works}
\label{s.relatedworks}

In Computer Vision, several works have presented methods to process
and classify images of parasites from different means (e.g., blood,
intestine, water, and skin). Most works rely on classical image
processing and machine learning
techniques~\cite{Yang:2001:BIOENG,Ginoris:2007,Castanon:2007,Dogantekin:2008:EXPERT,Avci:2009:EXPERT,Suzuki:2013:TBME,
  sengul2016classification}. Recent efforts have also
  been made towards the automated diagnosis of human intestinal
  parasites~\cite{alva2017mathematical,nkamgang2018expert,tchinda2018towards,NKAMGANG201881}.
  However, it is difficult to compare them with our work. They do not
  usually mention the parasitological protocol adopted to create the
  microscopy slides, which is crucial to make the automated image
  analysis feasible --- e.g., a considerable reduction in fecal
  impurities may facilitate the image analysis, but at the cost of
  losing some species of parasites in the microscopy slide. The DAPI
  system is a complete solution, from the collection, storage,
  transportation, and processing of fecal samples to create microscopy
  slides for automated image acquisition and analysis using a
  compromise focus plane. In those works, fecal impurities are not
  usually present and the images are acquired with manual focus. We
  then compare the methods in this paper with 13 deep neural networks
  and previous
  works~\cite{Suzuki:2013:TBME,Suzuki:2013:ISBI,peixinho2015diagnosis},
  all based on the DAPI system.

\begin{figure}[!ht]
    \begin{center}
   
    \begin{tabular}{cc}
      \includegraphics[height=4cm]{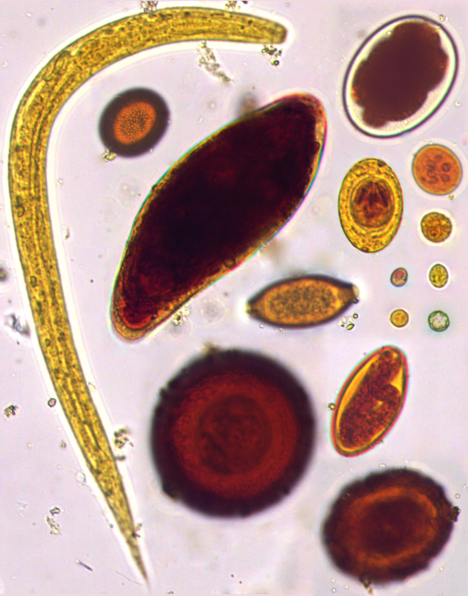} & \includegraphics[angle=90,height=4cm]{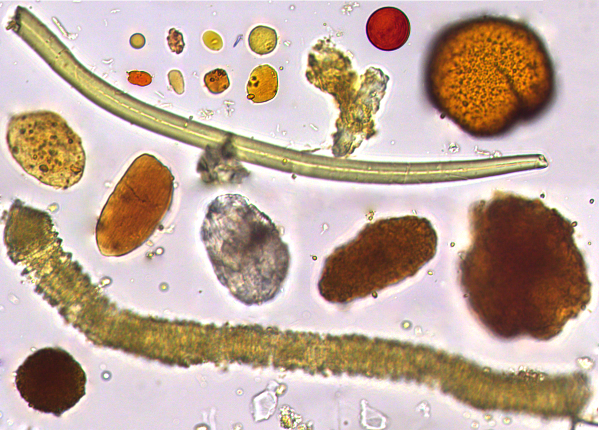} \\
      \rowcolor{white}
      (a) & (b)
      \end{tabular}
    \end{center}
  \caption{(a) The 15 most common species of human intestinal
    parasites in Brazil  and (b) similar impurity objects ~\cite{peixinho2015diagnosis}.
    \label{fig:all_parasites}}
\end{figure}

\begin{figure}[!ht]
    \begin{center}
    
    \begin{tabular}{cccc}
      \includegraphics[width=.15\linewidth]{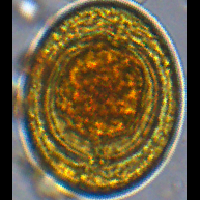} &
      \includegraphics[width=.15\linewidth]{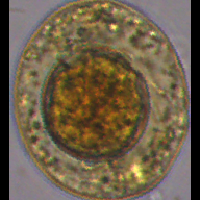} & 
      \includegraphics[width=.15\linewidth]{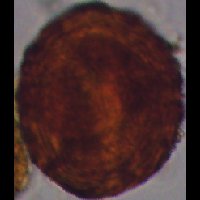} &
      \includegraphics[width=.15\linewidth]{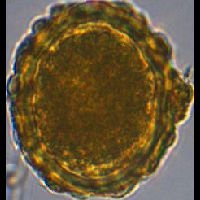}\\
      \rowcolor{white}
      (a) & (b) & (c) & (d)\\
      \includegraphics[width=.15\linewidth]{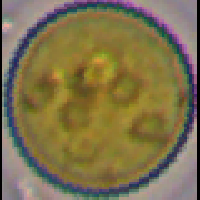} &
      \includegraphics[width=.15\linewidth]{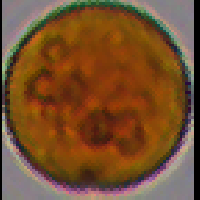} &
      \includegraphics[width=.15\linewidth]{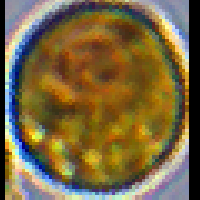} &
      \includegraphics[width=.15\linewidth]{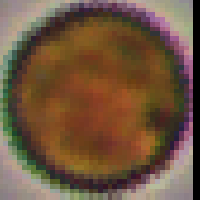}  \\
      (e) & (f) & (g) & (h)
      \end{tabular}
  \caption{Intra-class differences. (a-b) \emph{Hymenolepis nana}, (c-d) \emph{Ascaris lumbricoides}, (e-f) \emph{Entamoeba coli}, and (g-h) \textit{Iodamoeba b\"utschlii}.
    \label{fig:intraclass_variation}}
    \end{center}

\end{figure}

The DAPI system~\cite{Suzuki:2013:TBME} is composed
of one optical microscope with a motorized stage, focus driver, and
digital camera, all controlled by a
computer. In~\cite{Suzuki:2013:ISBI}, one can find improvements and
details about the image processing and machine learning techniques
adopted to find a common focus plane for image acquisition, to segment
objects (parasites and similar impurities) from the images, to align
the objects based on principal component analysis, to characterize
each object based on handcrafted (texture, color, and shape) features,
and to classify them into one out of 16 classes (15 species of
parasites and fecal impurity). In this work, we will use the same
preprocessing operations that separate the segmented objects into
three groups: (i) helminth eggs, (ii) protozoa cysts and vacuolar form
of protozoa, and (iii) helminth larvae, each with their similar fecal
impurities, for subsequent characterization and
classification. Figure~\ref{fig:all_parasites} illustrates how similar
can be these impurity objects to the real parasites and
Figure~\ref{fig:intraclass_variation} shows that even
  examples of the same class might be different among them, due to
  different living stages of the parasite.

Our preliminary works used a dataset with less than 8,000 images while
the present work uses a dataset with almost 52,000 images. A higher
number of images called our attention to the importance of using more
effective characterization and classification techniques --- a pair of
operations that we call a \emph{decision
  system}. In~\cite{Suzuki:2013:TBME}, for instance, the experiments
showed that the Optimum-Path Forest (OPF)
classifier~\cite{Papa:2009:IJIST} was the best choice for
classification. In this work, we show that $p$-SVM is a better choice
than OPF for our first decision system, $DS_1$.

\begin{sloppypar}

\begin{figure}[ht]
  \begin{center}
    \includegraphics[width=.75\columnwidth]{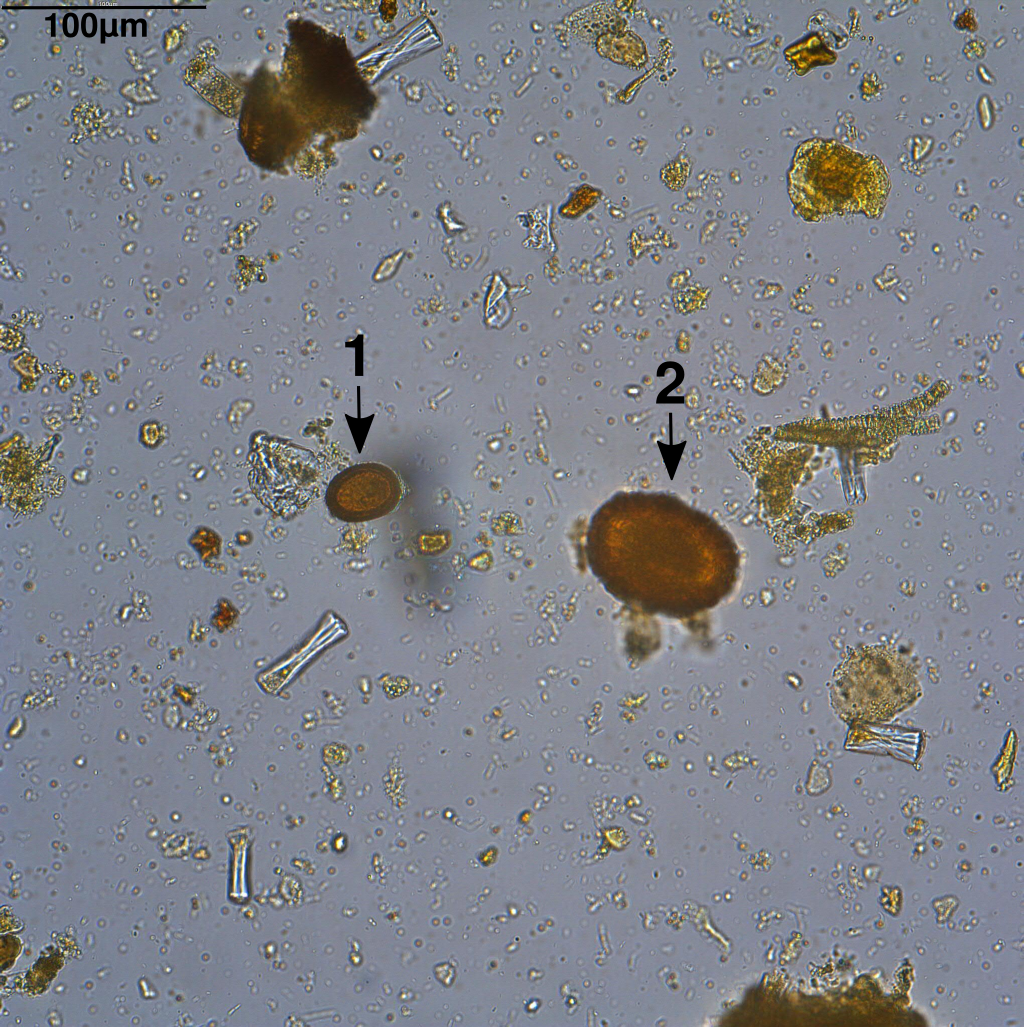}
    \caption{The input for object segmentation: field image from a microscopy slide showing a \emph{Taenia} spp. egg (Arrow 1) and an \emph{ A. lumbricoides} fertile egg (Arrow 2).  \label{fig:slide-field}}
  \end{center}
\end{figure}

In the meantime, we have investigated parasitological techniques to
create microscopy slides richer in parasites and with less
impurities~\cite{Carvalho2016TFTestMN} (see
Figure~\ref{fig:slide-field}), and we have also observed on a dataset
with 16,437 images that convolutional neural networks can considerably
improve characterization and classification of human intestinal
parasites~\cite{peixinho2015diagnosis}. Given that, we decided to
further investigate deep neural networks and came to the current
proposal of combining $DS_1$ with a decision system $DS_2$ based on
Vgg-16.
\end{sloppypar}

\section{Materials}
\label{s.materials}

\begin{sloppypar}In this section, we describe the datasets used for the experiments, which
contain a total of 51,919 images with the 15 most common species of human
intestinal parasites in Brazil and similar fecal impurities. Stool
samples have been obtained from the regions of Campinas and
Ara\c{c}atuba, S\~ao Paulo, Brazil, and processed in our lab
(Laboratory of Image Data Science/LIDS) in Campinas, S\~ao Paulo,
Brazil by using a parasitological technique called \textit{TF-Test
  Modified}~\cite{Carvalho2016TFTestMN} --- a technique based on the
centrifugation-sedimentation principle to concentrate parasites and
reduce the amount of fecal impurities in optical microscopy
slides. After parasitological processing, microscopy slides were
prepared and used in our system for automated image acquisition. The
objects in those images were automatically segmented, aligned, and
separated into three groups~\cite{Suzuki:2013:ISBI}, being the class
of each one identified by experts in Parasitology (or confirmed by the
experts after their automated recognition in our system). These groups
and classes are as follows.
\end{sloppypar}

\begin{enumerate}[(i)]
    \item \begin{sloppypar}Helminth eggs: \textit{H. nana, H. diminuta,}
      Ancylostomatidae \textit{, E. vermicularis, A. lumbricoides,
        T. trichiura, S. mansoni, Taenia} spp. and impurities of
      similar size and shape, named EGG-9;
      \end{sloppypar}
    \item Cysts and vacuolar form of protozoa: \textit{E. coli,
      E. histolytica / E. dispar, E. nana, Giardia, I. b\"utschlii,
      B. hominis} and impurities of similar size and shape, named
      PRO-7;
    \item Helminth larvae: \textit{S. stercoralis} and
      impurities with similar size and shape, named LAR-2;
\end{enumerate}

\begin{sloppypar}
Table \ref{table:parasites_database} presents the number of images in
our database per group and class in each group. One can notice that
each group represents one seriously unbalanced dataset with
considerably more fecal impurities than parasites. Indeed, this
reflects what is likely found in regular exams. Given that unbalanced
datasets can critically affect the performance of classification
systems and the impurities are numerous with similar examples to any
other category, this can be considered a challenging problem.
\end{sloppypar}
\begin{table}[!htb]

\begin{center}
\resizebox{0.45\textwidth}{!}{
\begin{tabular}{ |l|l|l|l| }

\hline
\multicolumn{4}{ |c| }{Parasites Database} \\
\hline
Database Groups & Number & Category & Class ID \\ \hline
\multirow{3}{*}{LAR-2} 
&         501     & \textit{Strongyloides stercoralis} & 1\\
&         1351    &         Impurities & 2\\
& \textbf{1852}   & \textbf{Total} &\\ \hline
 
\multirow{10}{*}{EGG-9}  
&         501    & \textit{Hymenolepis nana} & 1\\
&         83     & \textit{Hymenolepis diminuta }& 2\\
&         286    & Ancylostomatidae &3\\ 
&         103    & \textit{Enterobius vermicularis} &4\\ 
&         835    & \textit{Ascaris lumbricoides} &5\\ 
&         435    & \textit{Trichuris trichiura} &6\\ 
&         254    & \textit{Schistosoma mansoni} &7\\
&         379    & \textit{Taenia} spp. &8\\
&         9815   &         Impurities& 9\\
& \textbf{12691} & \textbf{Total} &\\ \hline

\multirow{8}{*}{PRO-7} 
&         869     & \textit{Entamoeba coli} &1\\
&         659     & \textit{Entamoeba histolytica / E. dispar} &2\\
&         1783    & \textit{Endolimax nana} &3\\
&         1931    & \textit{Giardia duodenalis} &4\\
&         3297    & \textit{Iodamoeba b\"utschlii} &5\\
&         309     & \textit{Blastocystis hominis} &6\\
&         28528   &         Impurities& 7\\
& \textbf{37376}  & \textbf{Total} &\\ \hline
      
\hline
\end{tabular}
}

\end{center}
\caption{Most common species of human intestinal parasites in Brazil: number of images per group and category in each group. \label{table:parasites_database}}
\end{table}

The experiments use a PC with the following
  specifications: Intel (R) Core (TM) i7-7700 - 3.60 GHz (8 CPU), RAM
  - 64 GiB, LINUX operating system (Ubuntu - 16:04 LTS - 64 bit) and
  GeForce GTX 1060 6GB/PCIe/SSE2.

As we will see next, our methodology requires a validation
set. Therefore, each group, EGG-9, PRO-7, and LAR-2, is divided into
training, validation, and testing sets for the experiments (as
described in Section~\ref{s.experiments}) and this process is also
repeated 10 times to obtain reliable statistical results.

\section{Methods}
\label{s.methods}

\begin{figure}[!ht]
  \begin{center}
    \includegraphics[width=.9\linewidth]{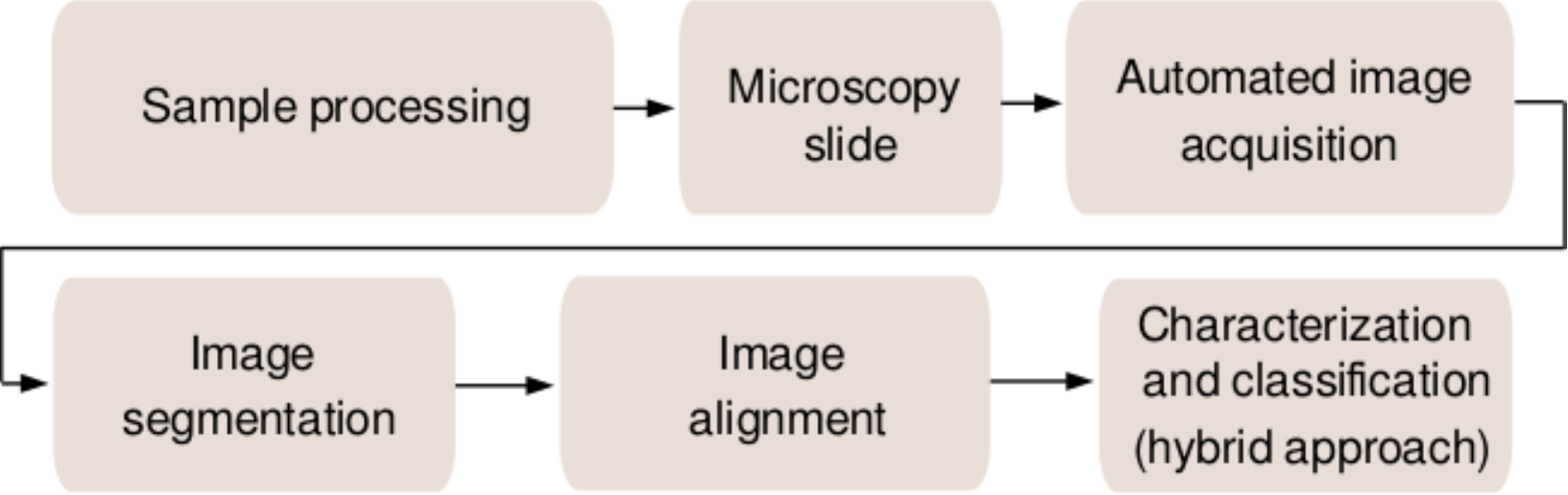} 
  \end{center}
  \caption{Data flow of the DAPI system. \label{f.DAPI-flowchart}}
\end{figure}

\begin{figure}[!ht]
  \begin{center}
    \includegraphics[width=.7\linewidth]{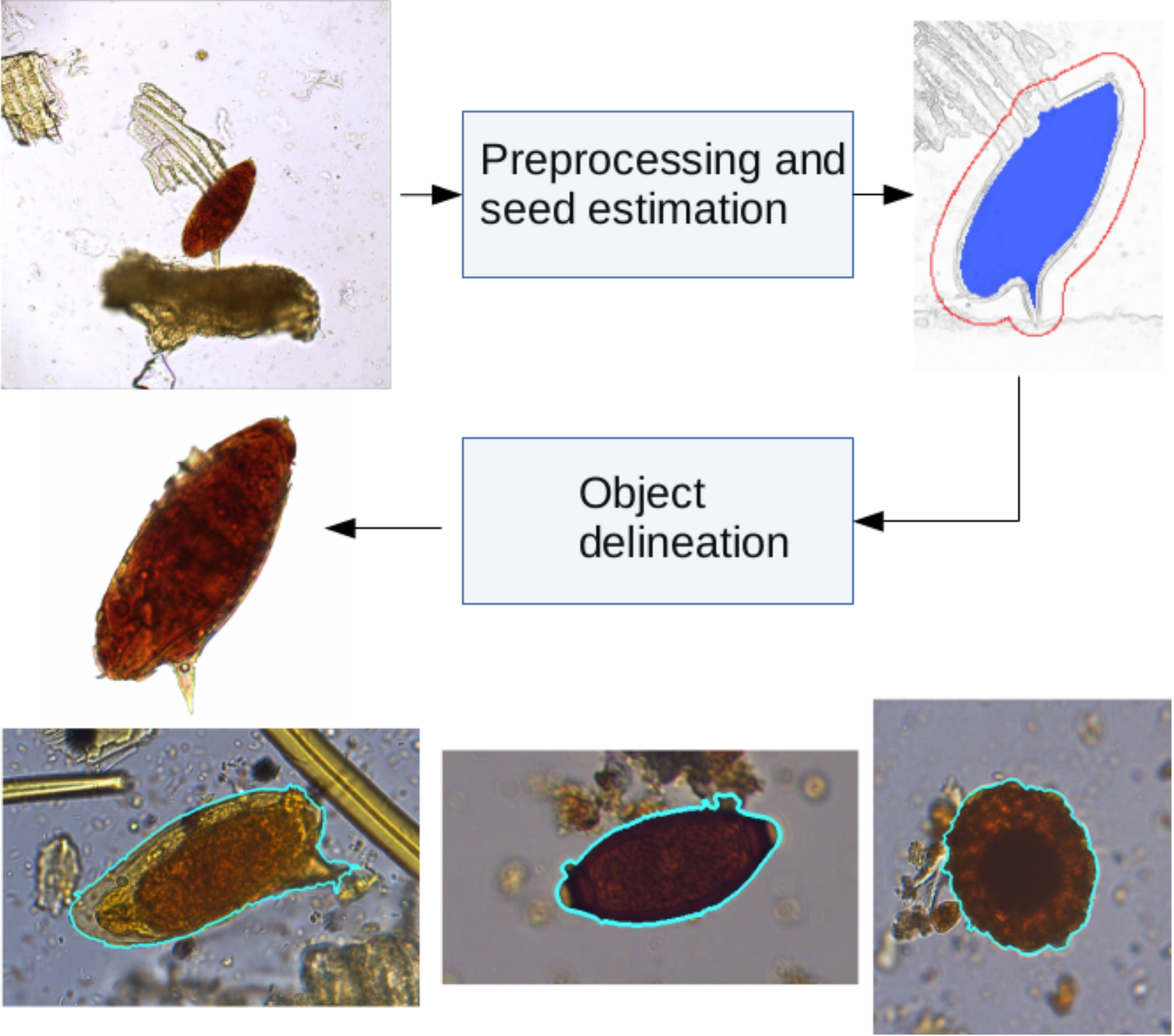} 
  \end{center}
  \caption{At the top, each slide field image is preprocessed by
    enhancing, automatic thresholding, IFT-based connected operators,
    and an area filter in order to estimate internal (blue) and
    external (red) seed pixels, as shown on a gradient image. Object
    delineation uses the IFT-watershed operator --- seeds compete
    among themselves and the object is defined by the pixels conquered
    by minimum-cost paths from the internal seeds, where the cost of a
    path is the maximum gradient value along it. At the bottom, it
    shows three other examples of segmentation (cyan) in the presence
    of impurities: \emph{S. mansoni} (left), \emph{T. trichiura}
    (center), and \emph{A. lumbricoides} (right).
  \label{f.segmentation}}
\end{figure}

Figure~\ref{f.DAPI-flowchart} shows a flow chart of
  the DAPI system. Fecal samples are collected, stored, transported by
  using the TF-Test kit and processed in our laboratory by using the
  TF-Test Modified protocol~\cite{Carvalho2016TFTestMN}, creating an
  optical microscopy slide for automated image acquisition by
  following a single compromise plane of
  focus~\cite{Suzuki:2013:TBME,Suzuki:2013:ISBI}. Each slide generates
  about 2,000 images (e.g., Figure~\ref{fig:slide-field}), which might
  be out of focus for some objects. Image segmentation involves a
  sequence of IFT-based image processing
  operations~\cite{Falcao:2004:TPAMI} suitable to separate parasites
  and impurities, in some situations that they appear connected, as
  illustrated in Figure~\ref{f.segmentation}. The segmentation mask is
  used for image alignment by principal component analysis --- i.e.,
  to align an image of a region of interest (ROI) around each object
  (Figure~\ref{fig:intraclass_variation}), which is used as input to
  the proposed hybrid decision system. 

For any group, EGG-9, PRO-7, or LAR-2, let $Z_1$, $Z_2$, and $Z_3$ be
its corresponding training, validation, and testing sets, $Z_1\cap Z_2
\cap Z_3=\emptyset$. Each image (after image segmentation and
alignment) $s\in Z_1 \cup Z_2 \cup Z_3$ may come from one of $m$
classes $w_j$, $j=1,2,\ldots,m$. The proposed hybrid system relies on
two decision-making systems with complementary properties:
\begin{itemize}
  \item $DS_1$ is based on very fast handcrafted feature
    extraction~\cite{Suzuki:2013:ISBI} and $p$-SVM
    classification~\cite{wu2004probability}.

  \item $DS_2$ is based on Vgg-16~\cite{vggNet} for feature extraction
    and classification. 
\end{itemize}
\begin{sloppypar}
For characterization in $DS_1$, the method uses the
  aligned segmentation mask and combines the BIC color descriptor
  from~\cite{Stehling-CIKM2001} with the object area, perimeter,
  symmetry, major and minor axes of the best fit ellipse within the
  object, the difference between the ellipse and the object, energy,
  entropy, variance, and homogeneity of a co-occurrence matrix. These
  features are used with different weights in a single feature
  vector. The Multi-Scale Parameter Search algorithm
  (MSPS)~\cite{Ruppert-CMBBE2017} is applied to find the weight of
  each feature that maximizes classification accuracy in the training
  set. The final weights modify the Euclidean distance between the
  feature vectors of two objects during classification. In $DS_2$,
  characterization is obtained at the output of the last convolutional
  layer of Vgg-16 (similarly to any other CNN, before the layers of a
  MLP classifier) and its parameters are learned by
  backpropagation. Note that, $DS_1$ depends on the success of
  segmentation more than $DS_2$. For $DS_2$, segmentation only affects
  the alignment of the input ROI image.
\end{sloppypar}

While $DS_1$ is about 30 times faster than $DS_2$, the latter is
considerably more accurate than the former. The main
  idea is that $DS_1$ should assign a class $w_j$, $j\in [1,m]$, to an
  image $s$, with a confidence value that $s$ comes from $w_j$. The
  hybrid method uses that confidence value to select the most likely
  misclassified images by $DS_1$ to be processed by $DS_2$. However,
  in order to improve effectiveness without compromising efficiency
  and cost, a limited number of images must be selected from the 2,000
  images of a microscopy slide.

Let $c$ be a random variable that represents the
  confidence values assigned to images by $DS_1$. The probability of
  error $P_{error}(c\backslash w_j)$, when $DS_1$ assigns an image $s$
  to a class $w_j$ with confidence value $c$, should be inversely
  proportional to $c$, but we have observed that this is not usually
  the case with $p$-SVM. In order to circumvent the problem, we use a
  validation set to estimate probability distributions
  $P_{error}(c_i\backslash w_j)$, $j=1,2,\ldots,m$, as normalized
  histograms with $n$ intervals (bins) $c_i$, $i=1,2,\ldots,n$. When
  $DS_1$ classifies the objects extracted from images of a microscopy
  slide, the object images with confidence values that fall in bins
  with higher probability of error $P_{error}(c_i\backslash w_j)$ have
  higher priority to be selected for characterization and
  reclassification by $DS_2$.

The proposed training and testing phases of $DS_1$ and $DS_2$ for the
hybrid method are described next.

\subsection{Training our hybrid decision system}
\label{ss.training}

\begin{figure}[ht]
  \begin{center}
    \begin{tabular}{cc}
      \includegraphics[width=.45\linewidth]{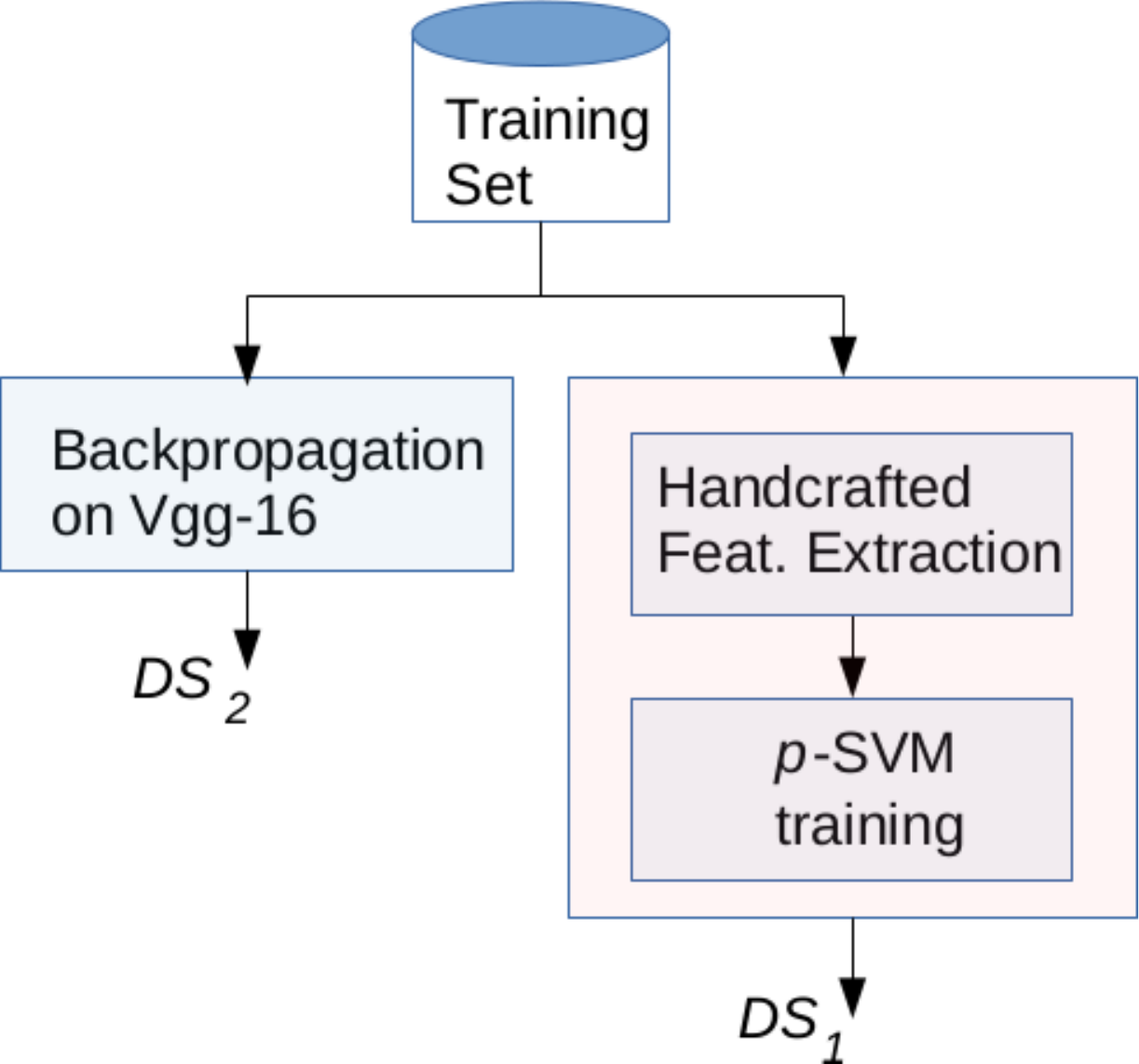} &
      \includegraphics[width=.45\linewidth]{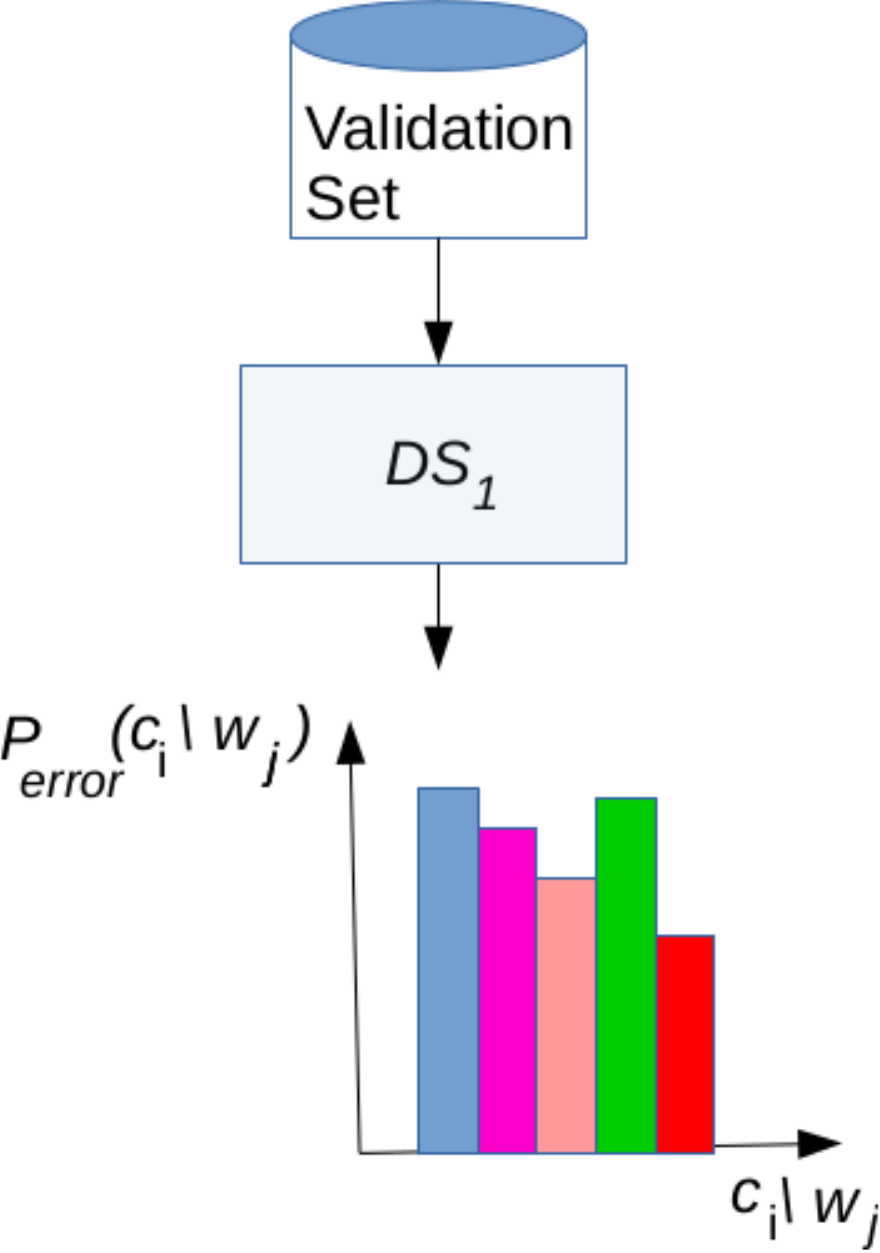} \\
      (a) &
      (b) 
    \end{tabular}
  \end{center}
  \caption{(a) Training of the decision systems $DS_1$ and $DS_2$. (b)
    After training, $DS_1$ is used on a validation set to estimate
    $P_{error}(c_i\backslash w_j)$ for each class $w_j$, $j=1,2,\ldots,m$,
    and interval (bin) $c_i$, $i=1,2,\ldots,n$.}
  \label{f.training}
\end{figure}

Figure~\ref{f.training}a illustrates the training
  processes of $DS_1$ and $DS_2$ using images from $Z_1$. While Vgg-16
  is trained by backpropagation from pre-aligned ROI
  images~\cite{vggNet}, $p$-SVM~\cite{wu2004probability} is trained
  from the handcrafted features extracted from pre-segmented and
  aligned objects, as proposed in~\cite{Suzuki:2013:ISBI}. The
  implementation details about both training processes are given in
  Section~\ref{ss.exper1}.

\begin{figure}
    \centerline{
    \includegraphics[width=9cm]{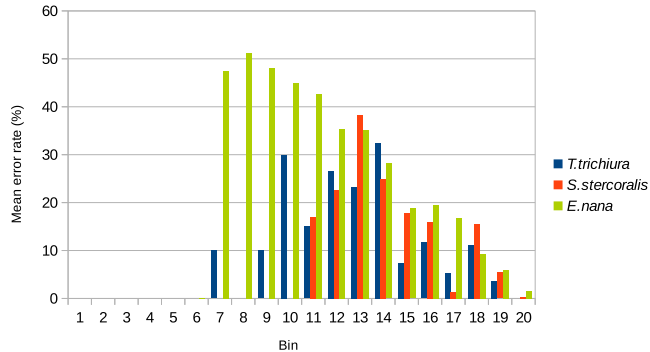} 
    }
    \caption{The histograms $P_{error}(c_i\backslash w_j)$ for three
      classes of helminth eggs and $n=20$ bins.}
    \label{f.OVO_probability}
\end{figure}

In Figure~\ref{f.training}b, we estimate
  $P_{error}(c_i\backslash w_j)$ by dividing the confidence values of
  $DS_1$ into $n > 1$ intervals $c_i$ (bins), $i=1,2,\ldots,n$, and
  computing for each bin the percentage of images from a validation
  set $Z_2$ that are misclassified by $DS_1$ as belonging to each
  class $w_j$, $j=1,2,\ldots,m$. This results into $m$ histograms
  $P_{error}(c_i\backslash w_j)$, as illustrated in
  Figure~\ref{f.OVO_probability} for three classes and $n=20$
  bins. Note that they are not inversely proportional to the
  confidence values for any class.

\subsection{Testing our hybrid decision system}
\label{ss.testing}

\begin{figure}[ht]
  \begin{center}
    \includegraphics[width=.6\linewidth]{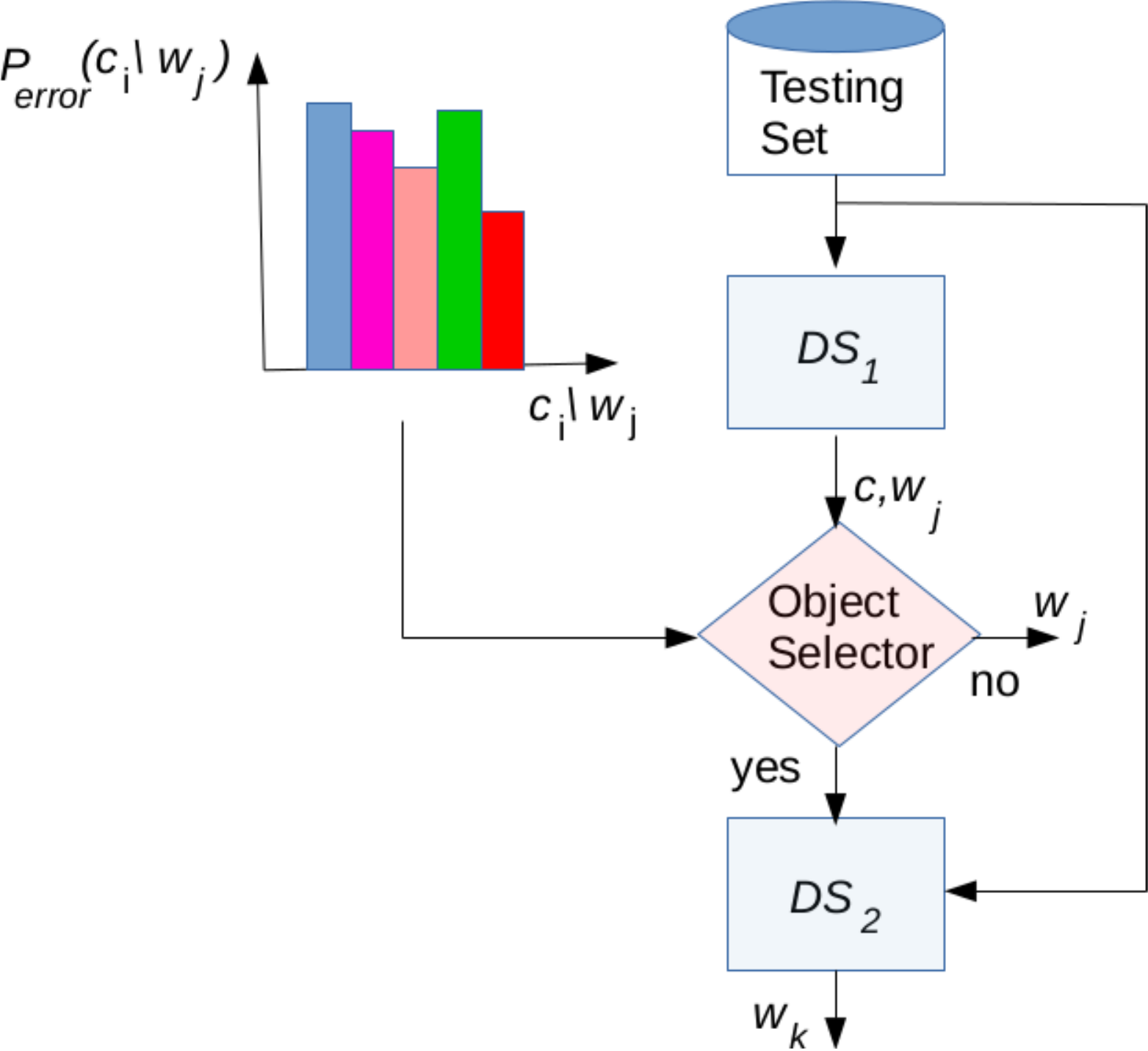} 
  \end{center}
  \caption{For each object in $Z_3$, $DS_1$ assigns a class $w_j$,
    $j\in [1,m]$, with confidence value $c$ that falls in a bin $c_i$,
    $i\in [1,n]$. Based on $P_{error}(c_i\backslash w_j)$, the
    selector decides which object images will be processed by
    $DS_2$. It randomly picks images in their decreasing order of
    $P_{error}(c_i\backslash w_j)$, until the total number of selected
    images is $M$. The selected images might be reassigned to a class
    $w_k\neq w_j$, $k\in [1,m]$, by $DS_2$, otherwise $w_j$ is the
    assigned class.}
  \label{f.testing}
\end{figure}

A test set $Z_3$ contains objects that are segmented
  for image alignment and feature extraction. $DS_1$ is used to
  classify those objects in one of the classes $w_j$, $j\in [1,m]$,
  with a confidence value $c$ that falls in one of the bins $c_i$,
  $i\in [1,n]$. A selector sorts the images of those objects by their
  decreasing order of $P_{error}(c_i\backslash w_j)$. For a given
  number $M$ of selected images for characterization and
  reclassification by $DS_2$, the selector randomly picks $M_j=M\times
  P_{error}(c_i\backslash w_j)$ images per class $w_j$,
  $j=1,2,\ldots,m$, and bin $c_i$, $i=1,2,\ldots,n$, by following
  their decreasing order of $P_{error}(c_i\backslash w_j)$, until the
  total number of selected images is $M$. The selected images might be
  reassigned to a class $w_k\neq w_j$, $k\in [1,m]$, by $DS_2$,
  otherwise $w_j$ is the assigned class (Figure~\ref{f.testing}).

\section{Experimental results and discussion}
\label{s.experiments}

For the experiments, each dataset, EGG-9, PRO-7, and LAR-2, is divided
into 40\% for training ($Z_1$), 30\% for validation ($Z_2$), and 30\%
for testing ($Z_3$) by using stratified random sampling. This process
is also repeated 10 times to obtain reliable statistical results.

First, we evaluate the dependence of our approach with respect to the
number $n$ of bins and select the best values for each group: $n=20$
bins for EGG-9 and $n=10$ bins for LAR-2 and PRO-7
(Section~\ref{ss.exper1}). We then show in the same section that our
approach can be considerably more accurate than $DS_1$ and almost so
accurate as $DS_2$ in most cases, being considerably faster than
$DS_2$ and almost so fast as $DS_1$ . We also show that a random
choice of object images for reclassification by $DS_2$ rather than a
choice based on $P_{error}(c_i\backslash w_j)$ is not a good option,
which reinforces the importance of our approach. Second, we justify
our choice for $p$-SVM in $DS_1$ by comparing different $SVM$ models
and the OPF classifier in Section~\ref{ss.exper2}. Finally, we justify
our choice for Vgg-16 in $DS_2$ by comparing it with several other
deep neural networks in Section~\ref{ss.exper3}.

\subsection{Hybrid approach versus $DS_1$ and $DS_2$}
\label{ss.exper1}

In $DS_1$, we optimize the parameters of $p$-SVM by using $Z1$ for
training and $Z_2$ for evaluation, with parameter optimization based
on grid search~\cite{Hsu10apractical}. In $DS_2$, we started from
Vgg-16 pre-trained on the Imagenet dataset and fine tuned it with the
training set $Z_1$. For fine tuning, we fixed the learning rate at
$1e^{-5}$, momentum at $0.9$, minibatch size at $16$, and the number
of epochs at $150$. As recommended for neural networks pre-trained
with the ImageNet dataset, the images were subtracted from the mean
values of each band and interpolated to $224 \times 224$ pixels and
$3$ bands.

Given that the hybrid approach depends on the number $n$ of bins for
$P_{error}(c_i\backslash w_j)$, we first show its performance on $Z_3$
using Vgg-16 in $DS_2$ and $p$-SVM in $DS_1$ with $n$ equal to 10, 20,
and 30 bins (Table~\ref{t.acc_bin}). One can see that the best number
of bins varies with the group: 10 bins for LAR-2 and PRO-7, and 20
bins for EGG-9. Therefore, these values are fixed for those groups in
the next experiment.

\begin{table}[!ht]
\resizebox{0.48\textwidth}{!}
{
\begin{tabular}{|l|l|l|l|l|}
\hline
 Dataset &  Technique   & 10 bins & 20 bins & 30 bins \\ \hline
EGG-9    &   proposed hybrid & 0.947  $\pm$ 0.005  &  \textbf{0.949  $\pm$  0.007} & 0.948 $\pm$ 0.005   \\ \hline
LAR-2    & proposed hybrid &  \textbf{0.894 $\pm$ 0.020}  &  0.878 $\pm$  0.031 & 0.869 $\pm$ 0.020   \\ \hline
PRO-7    & proposed hybrid & \textbf{0.926 $\pm$ 0.004}   &  0.925   $\pm$  0.004 & 0.925 $\pm$ 0.005  \\ \hline
\end{tabular}
}
\caption{Mean Cohen's Kappa and the respective standard deviation over
  each group of parasites using $n$ equal to 10, 20 and 30 bins to
  build $P_{error}(c_i\backslash w_j)$, $i\in [1,n]$ and $w_j\in
  [1,m]$.}
\label{t.acc_bin}
\end{table}

Tables~\ref{t.acc_perclass_egg},~\ref{t.acc_perclass_lar},
and~\ref{t.acc_perclass_pro} show the accuracy of each decision system
on $Z_3$ in each group, EGG-9, LAR-2, and PRO-7, respectively. We have
also added a variant of the hybrid system with random
selection (RS) of object images for characterization
and reclassification by $DS_2$. The hybrid approaches selected $M$
equal to 10\% of the images in $Z_3$ for $DS_2$. This number was
chosen to make the hybrid approaches acceptable for clinical routine
(e.g., 4 minutes per microscopy slide in the current computer
configuration).

\begin{table}
\resizebox{0.49\textwidth}{!}
{
\large
\begin{tabular}{|l|l|l|l|l|}
\hline
Class                 & \multicolumn{4}{c|}{Technique}               \\ \hline
                       & $DS_1$  & $DS_2$ & Hybrid with RS & Prop. hybrid \\ \hline
\emph{H.nana}          & 0.901  $\pm$ 0.020 &   0.995  $\pm$ 0.005 &   0.913  $\pm$ 0.016  & 0.987 $\pm$ 0.011          \\ \hline
\emph{H.diminuta}      & 0.736  $\pm$ 0.080 &   0.960  $\pm$ 0.018 &   0.780  $\pm$ 0.063  & 0.952 $\pm$ 0.025        \\ \hline
Ancylostomatidae       & 0.915  $\pm$ 0.030 &   0.986  $\pm$ 0.012 &   0.921  $\pm$ 0.030  & 0.984 $\pm$ 0.014        \\ \hline
\emph{E.vermicularis}  & 0.742  $\pm$ 0.106 &   0.984  $\pm$ 0.022 &   0.758  $\pm$ 0.093  & 0.981 $\pm$ 0.022        \\ \hline
\emph{A.lumbricoides}  & 0.739  $\pm$ 0.043 &   0.970  $\pm$ 0.018 &   0.753  $\pm$ 0.036  & 0.960 $\pm$ 0.014        \\ \hline
\emph{T.trichiura}     & 0.902  $\pm$ 0.024 &   0.993  $\pm$ 0.007 &   0.909  $\pm$ 0.020  & 0.979 $\pm$ 0.007        \\ \hline
\emph{S.mansoni}       & 0.649  $\pm$ 0.047 &   0.962  $\pm$ 0.022 &   0.670  $\pm$ 0.054  & 0.960 $\pm$ 0.021        \\ \hline
\emph{Taenia} spp.     & 0.803  $\pm$ 0.021 &   0.978  $\pm$ 0.014 &   0.821  $\pm$ 0.026  & 0.961 $\pm$ 0.018        \\ \hline
\emph{Impurities}      & 0.978  $\pm$ 0.003 &   0.994  $\pm$ 0.001 &   0.981  $\pm$ 0.002  & 0.982 $\pm$ 0.002        \\ \hline
\end{tabular}
}
\caption{Mean accuracy per class and the respective standard deviation
  over EGG-9 dataset, using $n=20$ bins in the proposed hybrid
  approach.}
\label{t.acc_perclass_egg}
\end{table}

\begin{table}
\resizebox{0.49\textwidth}{!}
{
\large
\begin{tabular}{|l|l|l|l|l|}
\hline
Class              & \multicolumn{4}{c|}{Technique}               \\ \hline
                     & $DS_1$            & $DS_2$ & Hybrid with RS & Prop. hybrid \\ \hline
\emph{S.stercoralis} & 0.771 $\pm$ 0.055  & 0.895  $\pm$ 0.040   &  0.776 $\pm$ 0.044  &  0.909  $\pm$  0.035  \\ \hline
\emph{Impurities}    & 0.979 $\pm$ 0.007  & 0.989  $\pm$ 0.005   &  0.979 $\pm$ 0.007  &  0.984  $\pm$  0.005  \\ \hline

\end{tabular}
}
\caption{Mean accuracy per class and the respective standard deviation over LAR-2 dataset, using $n=10$ bins in the proposed hybrid
  approach.}
\label{t.acc_perclass_lar}
\end{table}

\begin{table}
\resizebox{0.49\textwidth}{!}
{
\huge
\begin{tabular}{|l|l|l|l|l|}
\hline
Class               & \multicolumn{4}{c|}{Technique}               \\ \hline
                     & $DS_1$             & $DS_2$               & Hybrid with RS        & Prop. hybrid \\ \hline
\emph{E.coli}        & 0.897 $\pm$ 0.013  & 0.970  $\pm$ 0.010  &  0.907  $\pm$  0.012   &   0.967    $\pm$  0.013   \\ \hline
\emph{E.histolytica / E.  dispar} & 0.721 $\pm$ 0.030  & 0.878  $\pm$ 0.026  &  0.728  $\pm$  0.032   &   0.871    $\pm$  0.018   \\ \hline
\emph{E.nana}        & 0.847 $\pm$ 0.014  & 0.956  $\pm$ 0.018  &  0.857  $\pm$  0.016   &   0.952    $\pm$  0.016   \\ \hline
\emph{Giardia}       & 0.858 $\pm$ 0.017  & 0.965  $\pm$ 0.012  &  0.871  $\pm$  0.012   &   0.963    $\pm$  0.011   \\ \hline
\emph{I.b\"utschlii}   & 0.810 $\pm$ 0.018  & 0.957  $\pm$ 0.016  &  0.829  $\pm$  0.014   &   0.921    $\pm$  0.006   \\ \hline
\emph{B.hominis}     & 0.462 $\pm$ 0.040  & 0.723  $\pm$ 0.079  &  0.472  $\pm$  0.049   &   0.715    $\pm$  0.049   \\ \hline
\emph{Impurities}    & 0.977 $\pm$ 0.001  & 0.991  $\pm$ 0.002  &  0.980  $\pm$  0.001   &   0.982    $\pm$  0.001   \\ \hline
\end{tabular}
}
\caption{Mean accuracy per class and the respective standard deviation over PRO-7 dataset, using $n=10$ bins in the proposed hybrid
  approach.}
\label{t.acc_perclass_pro}
\end{table}

\begin{sloppypar}
One can see that the proposed hybrid approach can achieve competitive
performance with $DS_2$ in most cases (small differences of 1\% or
less, the exception is \emph{I.b\"utschlii} with about 3\% of difference
in accuracy). The comparison with the variant of the hybrid approach
with random object image selection also reveals that our method is
indeed able to select the images with higher chances of error by
$DS_1$ for characterization and reclassification by $DS_2$.
\end{sloppypar}

Table~\ref{t.execution_time} also shows the mean execution time in
milliseconds to characterize and classify a single image. From these
tables, one can see that the proposed hybrid approach can considerably
improve effectiveness with respect to $DS_1$ without compromising
efficiency and cost.

\begin{table}
\resizebox{0.48\textwidth}{!}
{
\begin{tabular}{|l|l|l|l|}
\hline
 Dataset &  $DS_1$ & $DS_2$ & Prop. hybrid \\ \hline
EGG-9 &  1.09  $\pm$ 0.009  &  16.77   $\pm$ 0.047  &  2.52 $\pm$ 0.011  \\ \hline

LAR-2 &  0.22  $\pm$ 0.019  &  35.45   $\pm$ 0.307  & 3.43 $\pm$ 0.029  \\ \hline

PRO-7 &  2.04 $\pm$ 0.018   &  14.55   $\pm$  0.079  & 3.18 $\pm$ 0.017 \\ \hline

\end{tabular}
}
\caption{Mean execution time in milliseconds and its standard
  deviation to characterize and classify one object image.}
\label{t.execution_time}
\end{table}

\subsection{Why have we chosen $p$-SVM for $DS_1$?}
\label{ss.exper2}

In principle, classifiers such as the Optimum-Path Forest (OPF)
classifier~\cite{Papa:2009:IJIST}, previously proposed for the
diagnosis of parasites in~\cite{Suzuki:2013:TBME}, and the SVM
models~\cite{wu2004probability}, $p$-SVM, SVM-OVA (one-versus-all),
and SVM-OVO (one-versus-one), can be modified to output a confidence
measure $c$. Therefore, any of them could have been used in
$DS_1$. However, $p$-SVM is the only one that directly outputs a
confidence measure $c$. OPF does not have parameters and the
parameters of the SVM models were optimized by grid search, as
described for $p$-SVM in Section~\ref{ss.exper1}. All classifiers were
trained on $Z_1$ and tested on $Z_3$.

\begin{sloppypar}
Table~\ref{t.svm_acc} presents the mean Cohen's Kappa among SVM-OVA,
SVM-OVO, $p$-SVM, and OPF on each group, EGG-9, LAR-2, and PRO-7. As
one can see, the SVM models are significantly more effective than OPF
on the current large datasets and, among the SVM models, $p$-SVM is
the most reasonable choice for $DS_1$ given its effectiveness and
direct confidence measure $c$.  
\end{sloppypar}

\begin{table}[!ht]
\resizebox{0.48\textwidth}{!}{
\begin{tabular}{|l|l|l|l|l|}
\hline
Dataset & \multicolumn{3}{c|}{SVM-based} & OPF \\ \hline
        & OVA     & OVO    & Probability &    \\ \hline
EGG-9   &  0.841 $\pm$ 0.012      & 0.840  $\pm$    0.014  &    0.840    $\pm$    0.013  & 0.699 $\pm$ 0.013 \\ \hline
LAR-2   &  0.772 $\pm$ 0.038      & 0.772  $\pm$    0.038  &    0.786    $\pm$    0.041  & 0.580 $\pm$ 0.033 \\ \hline
PRO-7   & 0.858  $\pm$ 0.006      & 0.847  $\pm$    0.006  &    0.847    $\pm$    0.005  & 0.635 $\pm$ 0.009 \\ \hline
\end{tabular}
}
\caption{Mean Cohen's Kappa obtained over each group using the SVM-based and OPF approaches.}
\label{t.svm_acc}
\end{table}

\subsection{Why have we chosen Vgg-16 for $DS_2$?}
\label{ss.exper3}

\begin{sloppypar}
We have compared $13$ deep neural networks (DNNs) pre-trained on the
Imagenet dataset and fine tuned with the pre-aligned training object
images in $Z_1$ in order to choose Vgg-16 for $DS_2$. These networks
have been trained as described in Section~\ref{ss.exper1} for
Vgg-16. However, the training process of these networks takes over one
month. In order to save time for the comparison among them, we have
adopted a different dataset partitioning here. We partitioned each
dataset such that 20\% of the images are used for the training set
$Z_1$ and 80\% of them are used for the testing set $Z_3$
($Z_2=\emptyset$), by stratified random sampling. We also repeated
this partitioning 10 times to obtain reliable statistical results.
\end{sloppypar}

Additionally, we have evaluated two options of training sets, with and
without a balanced number of images per class. In the balanced case,
we removed images such that the largest classes were represented by a
number of training samples equal to the size of the smallest class in
each group. This created almost balanced training sets to evaluate
their negative impact in classification. The test sets, however,
remained unbalanced in both cases.

Tables~\ref{tab:parasites_acc_egg}-~\ref{tab:parasites_acc_pro}
present the average results of accuracy and Cohen's Kappa on the
testing sets $Z_3$ of EGG-9, LAR-2, and PRO-7, respectively, using
balanced and unbalanced training sets. Clearly, the removal of
training samples to force balanced classes impairs the performances of
the DNNs. One may conclude that networks trained on
  balanced sets are not competitive with those trained on unbalanced
  sets and this should be expected, due to the fact that the test sets
  are unbalanced by nature. One may also conclude that Vgg-16 is among
  the best models in all groups, which justifies its choice for
  $DS_2$. Even in PRO-7 with unbalanced training sets, where
  Densenet-161 performed slightly better than Vgg-16, their difference
  of 0.001 in kappa, with a standard deviation 0.003, is not
  statistically significant. For the sake of efficiency and cost, it
  is also important to select the simplest model with the best overall
  effectiveness.

\begin{table}[!ht]
\centering
\resizebox{.48\textwidth}{!}{
\begin{tabular}{l|cccc}
Model & \multicolumn{2}{c}{balanced} & \multicolumn{2}{c}{unbalanced}\\
\hline 
 &  acc & kappa & acc & kappa\\
 \hline 
\textit{AlexNet}    &  $\mathbf{0.514}$ \tiny $\pm0.043$ & $\mathbf{0.308}$ \tiny $\pm0.029$ &  $\mathbf{0.936}$ \tiny $\pm0.004$ & $\mathbf{0.824}$ \tiny $\pm0.012$\\
\textit{Caffenet} &  $\mathbf{0.536}$ \tiny $\pm0.055$ & $\mathbf{0.346}$ \tiny $\pm0.039$ &  $\mathbf{0.973}$ \tiny $\pm0.002$ & $\mathbf{0.931}$ \tiny $\pm0.006$\\
\textit{Densenet-121} &  $\mathbf{0.484}$ \tiny $\pm0.047$ & $\mathbf{0.293}$ \tiny $\pm0.34$ &  $\mathbf{0.980}$ \tiny $\pm0.003$ & $\mathbf{0.951}$ \tiny $\pm0.004$\\
\textit{Densenet-161}   &  $\mathbf{0.528}$ \tiny $\pm0.045$ & $\mathbf{0.338}$ \tiny $\pm0.035$ &  $\mathbf{0.987}$ \tiny $\pm0.001$ & $\mathbf{0.966}$ \tiny $\pm0.003$\\
\textit{Densenet-169} &  $\mathbf{0.493}$ \tiny $\pm0.039$ & $\mathbf{0.307}$ \tiny $\pm0.027$ &  $\mathbf{0.984}$ \tiny $\pm0.002$ & $\mathbf{0.958}$ \tiny $\pm0.004$\\
\textit{GoogLenet} &  $\mathbf{0.613}$ \tiny $\pm0.052$ & $\mathbf{0.560}$ \tiny $\pm0.055$ &  $\mathbf{0.982}$ \tiny $\pm0.001$ & $\mathbf{0.960}$ \tiny $\pm0.004$\\
\textit{Inception-V3}   &  $\mathbf{0.316}$ \tiny $\pm0.121$ & $\mathbf{0.083}$ \tiny $\pm0.027$ &  $\mathbf{0.804}$ \tiny $\pm0.013$ & $\mathbf{0.432}$ \tiny $\pm0.044$\\
\textit{Resnet-50} &  $\mathbf{0.431}$ \tiny $\pm0.060$ & $\mathbf{0.265}$ \tiny $\pm0.041$ &  $\mathbf{0.976}$ \tiny $\pm0.010$ & $\mathbf{0.946}$ \tiny $\pm0.004$\\
\textit{Resnet101} &  $\mathbf{0.432}$ \tiny $\pm0.041$ & $\mathbf{0.254}$ \tiny $\pm0.022$ &  $\mathbf{0.977}$ \tiny $\pm0.001$ & $\mathbf{0.942}$ \tiny $\pm0.002$\\
\textit{Resnet-152}   &  $\mathbf{0.503}$ \tiny $\pm0.037$ & $\mathbf{0.306}$ \tiny $\pm0.028$ &  $\mathbf{0.983}$ \tiny $\pm0.002$ & $\mathbf{0.955}$ \tiny $\pm0.005$\\
\textit{Squeezenet} &  $\mathbf{0.547}$ \tiny $\pm0.083$ & $\mathbf{0.341}$ \tiny $\pm0.064$ &  $\mathbf{0.968}$ \tiny $\pm0.004$ & $\mathbf{0.908}$ \tiny $\pm0.032$\\
\textit{Vgg-16} &  $\mathbf{0.791}$ \tiny $\pm0.032$ & $\mathbf{0.610}$ \tiny $\pm0.042$ &  $\mathbf{0.989}$ \tiny $\pm0.001$ & $\mathbf{0.972}$ \tiny $\pm0.003$\\
\textit{Vgg-19}   &  $\mathbf{0.765}$ \tiny $\pm0.039$ & $\mathbf{0.577}$ \tiny $\pm0.048$ &  $\mathbf{0.988}$ \tiny $\pm0.001$ & $\mathbf{0.969}$ \tiny $\pm0.004$\\
\hline 
\end{tabular}
}
\caption{Comparison among DNNs over EGG-9 using balanced and unbalanced (stratified) training sets.}
\label{tab:parasites_acc_egg}
\end{table}

\begin{table}[!ht]
\centering
\resizebox{.48\textwidth}{!}{
\begin{tabular}{l|cccc}
Model & \multicolumn{2}{c}{balanced} & \multicolumn{2}{c}{unbalanced}\\
\hline 
 &  acc & kappa & acc & kappa\\
 \hline 
\textit{AlexNet}   &  $\mathbf{0.874}$ \tiny $\pm0.025$ & $\mathbf{0.618}$ \tiny $\pm0.055$ &  $\mathbf{0.941}$ \tiny $\pm0.005$ & $\mathbf{0.759}$ \tiny $\pm0.026$\\
\textit{Caffenet} &  $\mathbf{0.844}$ \tiny $\pm0.029$ & $\mathbf{0.549}$ \tiny $\pm0.049$ &  $\mathbf{0.925}$ \tiny $\pm0.006$ & $\mathbf{0.708}$ \tiny $\pm0.020$\\
\textit{Densenet-121} &  $\mathbf{0.837}$ \tiny $\pm0.036$ & $\mathbf{0.526}$ \tiny $\pm0.076$ &  $\mathbf{0.944}$ \tiny $\pm0.010$ & $\mathbf{0.759}$ \tiny $\pm0.045$\\
\textit{Densenet-161}   &  $\mathbf{0.869}$ \tiny $\pm0.039$ & $\mathbf{0.605}$ \tiny $\pm0.076$ &  $\mathbf{0.951}$ \tiny $\pm0.008$ & $\mathbf{0.792}$ \tiny $\pm0.039$\\
\textit{Densenet-169} &  $\mathbf{0.831}$ \tiny $\pm0.046$ & $\mathbf{0.521}$ \tiny $\pm0.085$ &  $\mathbf{0.945}$ \tiny $\pm0.006$ & $\mathbf{0.766}$ \tiny $\pm0.029$\\
\textit{Googlenet} &  $\mathbf{0.875}$ \tiny $\pm0.028$ & $\mathbf{0.702}$ \tiny $\pm0.049$ &  $\mathbf{0.948}$ \tiny $\pm0.004$ & $\mathbf{0.827}$ \tiny $\pm0.024$\\
\textit{Inception-V3}   &  $\mathbf{0.771}$ \tiny $\pm0.041$ & $\mathbf{0.153}$ \tiny $\pm0.104$ &  $\mathbf{0.498}$ \tiny $\pm0.166$ & $\mathbf{0.136}$ \tiny $\pm0.063$\\
\textit{Resnet-50} &  $\mathbf{0.800}$ \tiny $\pm0.031$ & $\mathbf{0.457}$ \tiny $\pm0.053$ &  $\mathbf{0.944}$ \tiny $\pm0.008$ & $\mathbf{0.763}$ \tiny $\pm0.039$\\
\textit{Resnet101} &  $\mathbf{0.806}$ \tiny $\pm0.052$ & $\mathbf{0.444}$ \tiny $\pm0.072$ &  $\mathbf{0.931}$ \tiny $\pm0.006$ & $\mathbf{0.702}$ \tiny $\pm0.031$\\
\textit{Resnet-152}   &  $\mathbf{0.831}$ \tiny $\pm0.047$ & $\mathbf{0.509}$ \tiny $\pm0.071$ &  $\mathbf{0.940}$ \tiny $\pm0.005$ & $\mathbf{0.746}$ \tiny $\pm0.023$\\
\textit{Squeezenet} &  $\mathbf{0.869}$ \tiny $\pm0.031$ & $\mathbf{0.595}$ \tiny $\pm0.063$ &  $\mathbf{0.937}$ \tiny $\pm0.008$ & $\mathbf{0.745}$ \tiny $\pm0.032$\\
\textit{Vgg-16} &  $\mathbf{0.898}$ \tiny $\pm0.029$ & $\mathbf{0.683}$ \tiny $\pm0.065$ &  $\mathbf{0.962}$ \tiny $\pm0.005$ & $\mathbf{0.848}$ \tiny $\pm0.020$\\
\textit{Vgg-19}   &  $\mathbf{0.896}$ \tiny $\pm0.018$ & $\mathbf{0.671}$ \tiny $\pm0.038$ &  $\mathbf{0.956}$ \tiny $\pm0.006$ & $\mathbf{0.830}$ \tiny $\pm0.022$\\
\hline 
\end{tabular}
}
\caption{Comparison among DNNs over LAR-2 using balanced and unbalanced training sets.}
\label{tab:parasites_acc_lar}
\end{table}

\begin{table}[!ht]
\centering
\resizebox{.48\textwidth}{!}{
\begin{tabular}{l|cccc}
Model & \multicolumn{2}{c}{balanced} & \multicolumn{2}{c}{unbalanced}\\
\hline 
 &  acc & kappa & acc & kappa\\
 \hline 
\textit{AlexNet}   &  $\mathbf{0.584}$ \tiny $\pm0.023$ & $\mathbf{0.356}$ \tiny $\pm0.018$ &  $\mathbf{0.942}$ \tiny $\pm0.002$ & $\mathbf{0.852}$ \tiny $\pm0.007$\\
\textit{Caffenet} &  $\mathbf{0.636}$ \tiny $\pm0.018$ & $\mathbf{0.421}$ \tiny $\pm0.016$ &  $\mathbf{0.966}$ \tiny $\pm0.002$ & $\mathbf{0.916}$ \tiny $\pm0.005$\\
\textit{Densenet-121} &  $\mathbf{0.479}$ \tiny $\pm0.027$ & $\mathbf{0.271}$ \tiny $\pm0.017$ &  $\mathbf{0.961}$ \tiny $\pm0.001$ & $\mathbf{0.903}$ \tiny $\pm0.004$\\
\textit{Densenet-161}   &  $\mathbf{0.542}$ \tiny $\pm0.035$ & $\mathbf{0.327}$ \tiny $\pm0.029$ &  $\mathbf{0.971}$ \tiny $\pm0.001$ & $\mathbf{0.927}$ \tiny $\pm0.003$\\
\textit{Densenet-169} &  $\mathbf{0.515}$ \tiny $\pm0.045$ & $\mathbf{0.304}$ \tiny $\pm0.032$ &  $\mathbf{0.966}$ \tiny $\pm0.001$ & $\mathbf{0.916}$ \tiny $\pm0.002$\\
\textit{Googlenet} &  $\mathbf{0.561}$ \tiny $\pm0.025$ & $\mathbf{0.393}$ \tiny $\pm0.019$ &  $\mathbf{0.965}$ \tiny $\pm0.001$ & $\mathbf{0.918}$ \tiny $\pm0.004$\\
\textit{Inception-V3}   &  $\mathbf{0.248}$ \tiny $\pm0.031$ & $\mathbf{0.102}$ \tiny $\pm0.014$ &  $\mathbf{0.901}$ \tiny $\pm0.009$ & $\mathbf{0.748}$ \tiny $\pm0.022$\\
\textit{Resnet-50} &  $\mathbf{0.500}$ \tiny $\pm0.037$ & $\mathbf{0.288}$ \tiny $\pm0.026$ &  $\mathbf{0.958}$ \tiny $\pm0.017$ & $\mathbf{0.909}$ \tiny $\pm0.002$\\
\textit{Resnet101} &  $\mathbf{0.480}$ \tiny $\pm0.034$ & $\mathbf{0.273}$ \tiny $\pm0.024$ &  $\mathbf{0.960}$ \tiny $\pm0.001$ & $\mathbf{0.899}$ \tiny $\pm0.003$\\
\textit{Resnet-152}   &  $\mathbf{0.576}$ \tiny $\pm0.039$ & $\mathbf{0.351}$ \tiny $\pm0.034$ &  $\mathbf{0.970}$ \tiny $\pm0.002$ & $\mathbf{0.924}$ \tiny $\pm0.005$\\
\textit{Squeezenet} &  $\mathbf{0.520}$ \tiny $\pm0.032$ & $\mathbf{0.302}$ \tiny $\pm0.025$ &  $\mathbf{0.947}$ \tiny $\pm0.003$ & $\mathbf{0.868}$ \tiny $\pm0.008$\\
\textit{Vgg-16} &  $\mathbf{0.666}$ \tiny $\pm0.035$ & $\mathbf{0.451}$ \tiny $\pm0.033$ &  $\mathbf{0.970}$ \tiny $\pm0.001$ & $\mathbf{0.926}$ \tiny $\pm0.003$\\
\textit{Vgg-19}   &  $\mathbf{0.698}$ \tiny $\pm0.029$ & $\mathbf{0.485}$ \tiny $\pm0.030$ &  $\mathbf{0.970}$ \tiny $\pm0.002$ & $\mathbf{0.925}$ \tiny $\pm0.005$\\
\hline 
\end{tabular}
}
\caption{Comparison among DNNs over PRO-7 using balanced and unbalanced training sets.}
\label{tab:parasites_acc_pro}
\end{table}

\begin{figure}[ht]
  \begin{center}
    \begin{tabular}{ccc}
      \includegraphics[width=.13\linewidth]{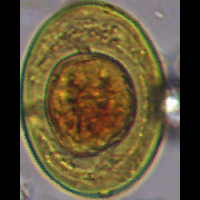} &
      \includegraphics[width=.13\linewidth]{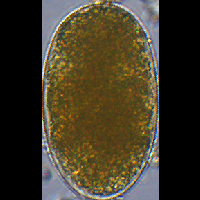} & 
      \includegraphics[width=.13\linewidth]{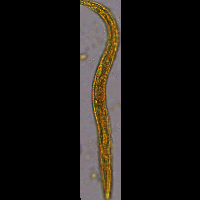}  \\ 
      (a) \emph{H. Nana} &   (b) \footnotesize Ancylostomatidae & (c)\emph{S. stercoralis} \\
      \includegraphics[width=.13\linewidth]{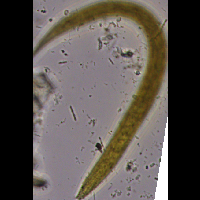} &
      \includegraphics[width=.13\linewidth]{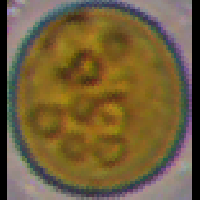} &
      \includegraphics[width=.13\linewidth]{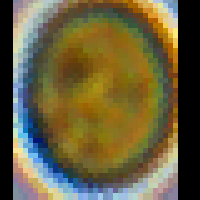} \\
       (d) \emph{S. stercoralis} & (e) \emph{E. coli}& (f) \emph{E. nana} \\

      \includegraphics[width=.13\linewidth]{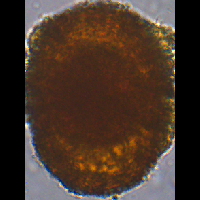} &
      \includegraphics[width=.13\linewidth]{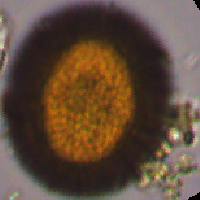} & 
      \includegraphics[width=.13\linewidth]{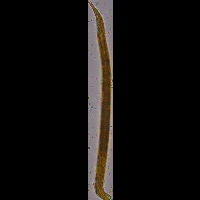}  \\ 
      (g) \footnotesize \emph{A. lumbricoides} &   (h) \emph{Taenia spp.} & (i) \emph{S. stercoralis} \\
      \includegraphics[width=.13\linewidth]{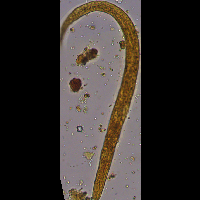} &
      \includegraphics[width=.13\linewidth]{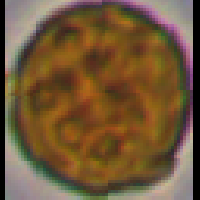} &
      \includegraphics[width=.13\linewidth]{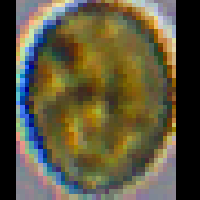} \\
      (j) \emph{S. stercoralis} & (k)\emph{E. coli} & (l) \emph{G. duodenalis}\\

    \end{tabular}
  \end{center}
  \caption{Examples of images (a-f) misclassified by $DS_1$, selected, and correctly classified by $DS_2$, (g-l) correctly classified by $DS_1$, selected, and misclassified by $DS_2$.}
  \label{f.examples}
\end{figure}

In summary, the justification for our hybrid approach is related to
two cases, when images are selected for reclassification by $DS_2$:
\begin{enumerate}
  \item \label{case1} For images that have been correctly classified by
    $DS_1$, the percentage of misclassifications by $DS_2$ should be low.
  \item \label{case2} For images that have been misclassified by
    $DS_1$, the percentage of correct classifications by $DS_2$ should
    be high.
\end{enumerate}
\begin{sloppypar}
Indeed, we have observed that, for the images that fall in
case~\ref{case1}, the percentage of misclassifications by $DS_2$ is
only $0.43$\% in EGG-9, $3.4$\% in PRO-7, and $3.22\%$ in LAR-2. For the
images that fall in case~\ref{case2}, the percentage of correct
classifications by $DS_2$ is $98.65$\% in EGG-9, $95.06$\% in PRO-7,
and $77.77\%$ in LAR-2. Examples of both cases are presented in
Figure~\ref{f.examples}.  
\end{sloppypar}

\section{Conclusion}
\label{s.conclusions}

\begin{sloppypar}
We presented a hybrid approach to combine two decision-making systems
with complementary properties --- a faster and less accurate decision
system $DS_1$ with a slower and more accurate decision system $DS_2$
--- in order to improve overall effectiveness without compromising
efficiency and cost in image analysis. We have successfully
demonstrated this approach for the diagnosis of the 15 most common
species of human intestinal parasites in Brazil. In this application,
after exhaustive experiments, the best choices of classifiers for
$DS_1$ and $DS_2$ were $p$-SVM and Vgg-16. The resulting hybrid system
also represents a low-cost solution viable for the clinical routine,
which makes our contribution relevant for the Public Health system in
Brazil.
\end{sloppypar}

\begin{sloppypar}
Our main technical contribution is a method that
  learns the probability distributions of error per class, given the
  confidence values of $DS_1$ on a validation set during training, to
  select test samples with higher chances of error for a final
  characterization and reclassification by $DS_2$. We believe this
hybrid approach can be useful in several other real applications.
\end{sloppypar}

As future work, we intend to investigate methods that can simplify a
deep neural network without losing effectiveness, or design a
lightweight deep neural network with higher effectiveness. By that, we
aim at increasing the number of samples selected for $DS_2$, further
improving the capability of our hybrid system to process more images
per time.

\section{Acknowledgments}

\begin{sloppypar}
The authors thank FAPESP (Proc. 2017/12974-0 and 2014/12236-1), Immunocamp Ci\^encia e Tecnologia LTDA, and CNPq (Proc. 303808/2018-7).
\end{sloppypar} 


\bibliographystyle{unsrt}
\bibliography{references}

\begin{thebibliography}{10}

\bibitem{WHO:2018}
WHO~Fact Sheets.
\newblock Soil-transmitted helminth infections.
\newblock
  \url{http://www.who.int/news-room/fact-sheets/detail/soil-transmitted-helminth-infections},
  2018.
\newblock [Online; accessed October 24, 2018].

\bibitem{Carvalho2016TFTestMN}
J.~B. de~Carvalho, B.~M.~Dos Santos, J.~F. Gomes, C.~T.~N. Suzuki, S.~H.
  Shimizu, A.~X. Falc{\~a}o, J.~C. Pierucci, L.~V.~S. de~Matos, and K.~D.~S.
  Bresciani.
\newblock Tf-test modified: New diagnostic tool for human enteroparasitosis.
\newblock {\em Journal of clinical laboratory analysis}, 30 4:293--300, 2016.

\bibitem{Suzuki:2013:TBME}
C.~T.~N. Suzuki, J.~F. Gomes, A.~X. Falc{\~{a}}o, J.~P. Papa, and
  S.~Hoshino-Shimizu.
\newblock Automatic segmentation and classification of human intestinal
  parasites from microscopy images.
\newblock {\em IEEE Transactions on Biomedical Engineering}, 60(3):803--812,
  March 2013.

\bibitem{Suzuki:2013:ISBI}
C.~T.~N. Suzuki, J.~F. Gomes, A.~X. Falc\~{a}o, S.~H. Shimizu, and J.~P. Papa.
\newblock Automated diagnosis of human intestinal parasites using optical
  microscopy images.
\newblock In {\em {IEEE} International Symposium on Biomedical Imaging}, pages
  460--463, 2013b.

\bibitem{wu2004probability}
Ting-Fan Wu, Chih-Jen Lin, and R.~C Weng.
\newblock Probability estimates for multi-class classification by pairwise
  coupling.
\newblock {\em Journal of Machine Learning Research}, 5(Aug):975--1005, 2004.

\bibitem{vggNet}
K.~Simonyan and A.~Zisserman.
\newblock Very deep convolutional networks for large-scale image recognition.
\newblock {\em CoRR}, abs/1409.1556, 2014.

\bibitem{Yang:2001:BIOENG}
Y.~Seok Yang, D.~K. Park, H.~C. Kim, M.-H. Choi, and J.-Y. Chai.
\newblock Automatic identification of human helminth eggs on microscopic fecal
  specimens using digital image processing and an artificial neural network.
\newblock {\em Biomedical Engineering, IEEE Transactions on}, 48(6):718--730,
  2001.

\bibitem{Ginoris:2007}
Y.~P. Ginoris, A.~L. Amaral, A.~Nicolau, M.~A.~Z. Coelho, and Ferreira~E. C.
\newblock Development of an image analysis procedure for identifying protozoa
  and metazoa typical of activated sludge system.
\newblock {\em Water Research}, 41:2581--2589, 2007.

\bibitem{Castanon:2007}
C.~A.~B. Casta{\~n}{\'o}n, J.~S. Fraga, S.~Fernandez, A.~Gruber, and L.~F.
  Costa.
\newblock Biological shape characterization for automatic image recognition and
  diagnosis of protozoan parasites of the genus eimeria.
\newblock {\em Pattern Recognition}, 40:1899--1910, 2007.

\bibitem{Dogantekin:2008:EXPERT}
E.~Dogantekin, M.~Yilmaz, A.~Dogantekin, E.~Avci, and A.~Sengur.
\newblock A robust technique based on invariant moments - anfis for recognition
  of human parasite eggs in microscopic images.
\newblock {\em Expert Systems with Applications}, 35(3):728--738, 2008.

\bibitem{Avci:2009:EXPERT}
D.~Avci and A.~Varol.
\newblock An expert diagnosis system for classification of human parasite eggs
  based on multi-class svm.
\newblock {\em Expert Systems with Applications}, 36(1):43--48, 2009.

\bibitem{sengul2016classification}
G{\"o}khan Seng{\"u}l.
\newblock Classification of parasite egg cells using gray level cooccurence
  matrix and knn., 2016.

\bibitem{alva2017mathematical}
Alicia Alva, Carla Cangalaya, Miguel Quiliano, Casey Krebs, Robert~H. Gilman,
  Patricia Sheen, and Mirko Zimic.
\newblock Mathematical algorithm for the automatic recognition of intestinal
  parasites.
\newblock {\em PloS one}, 12(4):e0175646, 2017.

\bibitem{nkamgang2018expert}
O.~T. Nkamgang, D~Tchiotsop, H.~B. Fotsin, and P.~K. Talla.
\newblock An expert system for assistance in human intestinal parasitosis
  diagnosis.
\newblock {\em Biosens Bioelectron Open Acc: BBOA-128.}, 10:2577--2260, 2018.

\bibitem{tchinda2018towards}
Beaudelaire~Saha Tchinda, Michel Noubom, Daniel Tchiotsop, Valerie Louis-Dorr,
  and Didier Wolf.
\newblock Towards an automated medical diagnosis system for intestinal
  parasitosis.
\newblock {\em Informatics in Medicine Unlocked}, 13:101--111, 2018.

\bibitem{NKAMGANG201881}
Oscar~Takam Nkamgang, Daniel Tchiotsop, Beaudelaire~Saha Tchinda, and
  Hilaire~Bertrand Fotsin.
\newblock A neuro-fuzzy system for automated detection and classification of
  human intestinal parasites.
\newblock {\em Informatics in Medicine Unlocked}, 13:81 -- 91, 2018.

\bibitem{peixinho2015diagnosis}
A.~Z. Peixinho, S.~B. Martins, J.~E. Vargas, A.~X. Falc{\~a}o, J.~F. Gomes, and
  C.~T.~N. Suzuki.
\newblock Diagnosis of human intestinal parasites by deep learning.
\newblock In {\em Computational Vision and Medical Image Processing V:
  Proceedings of the 5th Eccomas Thematic Conference on Computational Vision
  and Medical Image Processing (VipIMAGE 2015, Tenerife, Spain, October 19-21,
  2015)}, page 107. CRC Press, 2015.

\bibitem{Papa:2009:IJIST}
J.~P. Papa, A.~X. Falc\~{a}o, and C.~T.~N. Suzuki.
\newblock Supervised pattern classification based on optimum-path forest.
\newblock {\em International Journal of Imaging Systems and Technology},
  19(2):120--131, 2009.

\bibitem{Falcao:2004:TPAMI}
A.~X. Falc{\~{a}}o, J.~Stolfi, and R.~de~Alencar~Lotufo.
\newblock The image foresting transform: theory, algorithms, and applications.
\newblock {\em IEEE Transactions on Pattern Analysis and Machine Intelligence},
  26(1):19--29, Jan 2004.

\bibitem{Stehling-CIKM2001}
R.~O. Stehling, M.~A. Nascimento, and A.~X. Falc{\~{a}}o.
\newblock A compact and efficient image retrieval approach based on
  border/interior pixel classification.
\newblock In {\em Proc. of the 11th Intl. Conf. on Information and Knowledge
  Management (CIKM)}, pages 102--109, McLean, VA, 2002.

\bibitem{Ruppert-CMBBE2017}
G.~C.~S. Ruppert, G.~Chiachia, F.~P.~G. Bergo, F.~O. Favretto, C.~L. Yasuda,
  A.~Rocha, and A.~X.~Falc\ {a}o.
\newblock Medical image registration based on watershed transform from
  greyscale marker and multi-scale parameter search.
\newblock {\em Computer Methods in Biomechanics and Biomedical Engineering:
  Imaging \& Visualization}, 5(2):138--156, 2017.

\bibitem{Hsu10apractical}
Chih wei Hsu, Chih chung Chang, and Chih jen Lin.
\newblock A practical guide to support vector classification, 2010.

\end{thebibliography}

\end{document}